\documentclass[lettersize,journal]{IEEEtran}
\usepackage{amsmath,amsfonts}
\usepackage{algorithmic}
\usepackage{algorithm}
\usepackage{array}
\usepackage{textcomp}
\usepackage{stfloats}
\usepackage{url}
\usepackage{verbatim}
\usepackage{graphicx}
\usepackage{cite}
\usepackage{multirow}
\usepackage{rotating}
\usepackage{hyperref}
\usepackage{bbding} 
\usepackage{bm}  
\usepackage{booktabs} 
\usepackage{subcaption}
\usepackage{makecell} 
\begin{document}

\title{Text-Derived Relational Graph-Enhanced Network for Skeleton-Based Action Segmentation}

\author{Haoyu~Ji,
        Bowen~Chen,
        Weihong~Ren,
        Wenze~Huang,
        Zhihao Yang,    
	Zhiyong Wang,~\IEEEmembership{Member,~IEEE}, and~Honghai~Liu$^*$,~\IEEEmembership{Fellow,~IEEE}%
\thanks{* corresponding author

This work is supported by the National Key Research and Development Program of China under Grant 2022YFB4700202, by the National Natural Science Foundation of China under Grant 62261160652, 52275013, 62206075 and 61733011, by the Shenzhen Science and Technology Program under Grant JCYJ20240813105137049.

Haoyu Ji, Bowen Chen, Weihong Ren, Wenze~Huang, Zhihao Yang, Zhiyong Wang, and Honghai Liu are with the State Key Laboratory of Robotics and Systems, Harbin Institute of Technology Shenzhen, Shenzhen 518055, China (e-mail:  jihaoyu1224@gmail.com, honghai.liu@icloud.com)

The code is available at \url{https://github.com/HaoyuJi/TRG-Net}.}}

\markboth{Journal of \LaTeX\ Class Files,~Vol.~14, No.~8, August~2021}%
{Shell \MakeLowercase{\textit{et al.}}: A Sample Article Using IEEEtran.cls for IEEE Journals}


\maketitle

\begin{abstract}
Skeleton-based Temporal Action Segmentation (STAS) aims to segment and recognize various actions from long, untrimmed sequences of human skeletal movements. Current STAS methods typically employ spatio-temporal modeling to establish dependencies among joints as well as frames, and utilize one-hot encoding with cross-entropy loss for frame-wise classification supervision. However, these methods overlook the intrinsic correlations among joints and actions within skeletal features, leading to a limited understanding of human movements. To address this, we propose a Text-Derived Relational Graph-Enhanced Network (TRG-Net) that leverages prior graphs generated by Large Language Models (LLM) to enhance both modeling and supervision. For modeling, the Dynamic Spatio-Temporal Fusion Modeling (DSFM) method incorporates Text-Derived Joint Graphs (TJG) with channel- and frame-level dynamic adaptation to effectively model spatial relations, while integrating spatio-temporal core features during temporal modeling. For supervision, the Absolute-Relative Inter-Class Supervision (ARIS) method employs contrastive learning between action features and text embeddings to regularize the absolute class distributions, and utilizes Text-Derived Action Graphs (TAG) to capture the relative inter-class relationships among action features. Additionally, we propose a Spatial-Aware Enhancement Processing (SAEP) method, which incorporates random joint occlusion and axial rotation to enhance spatial generalization. Performance evaluations on four public datasets demonstrate that TRG-Net achieves state-of-the-art results.

\end{abstract}


\section{Introduction}
\IEEEPARstart{T}{emporal} Action Segmentation (TAS), an advanced task in video understanding, aims to segment and recognize each action within long, untrimmed video sequences of human activities~\cite{TAS}. 
Similar to how semantic segmentation predicts labels for each pixel in an image, TAS predicts action labels for each frame in a video. As a significant task in computer vision, TAS finds applications in various domains such as medical rehabilitation, ~\cite{medical}, industrial monitoring~\cite{Openpack}, and activity analysis~\cite{sports}.

\begin{figure}[t]
  \centering
   \includegraphics[width=0.95\linewidth]{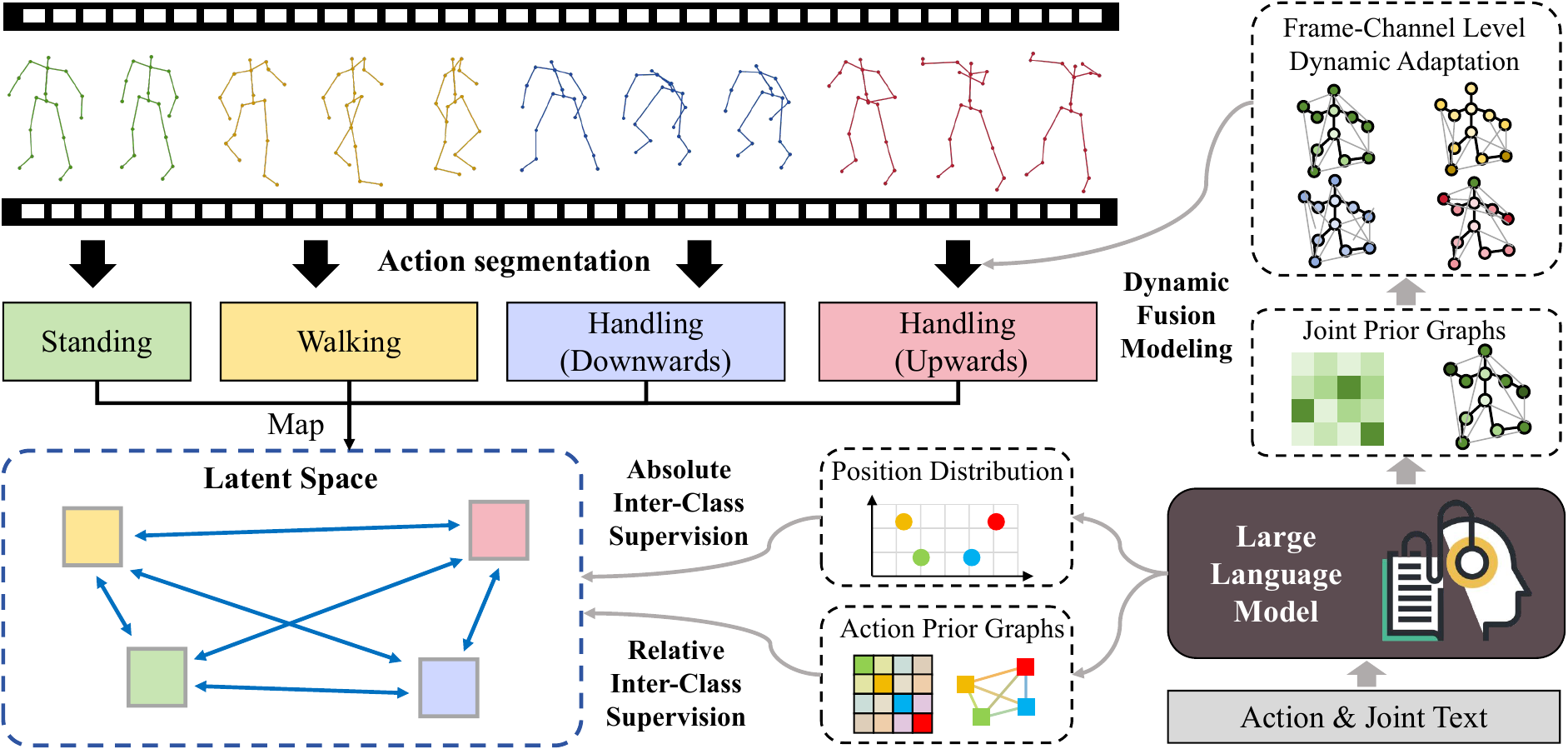}

   \caption{Schematic of TRG-Net concept. 
 The text embeddings and relational graphs generated by large language models can serve as priors for enhancing modeling and supervision of action segmentation. Specifically, the text-derived joint graph effectively captures spatial correlations, while the text-derived action graph and action embeddings supervise the relationships and distributions of action classes.}
   \label{fig:1}
\end{figure}

Existing TAS methods can be broadly categorized into two types based on input modality: Video-based TAS (VTAS) and Skeleton-based TAS (STAS)~\cite{IDT-GCN,CTC,LaSA}. VTAS uses input features such as RGB or optical flow~\cite{optical_flow2} extracted from video. While these features contain comprehensive information, they often suffer from redundancy, noise, and increased complexity in model fitting. In contrast, STAS, the field of this work, utilizes 2D or 3D human skeleton sequences~\cite{pose} extracted via motion capture devices or algorithms. STAS focuses exclusively on human motion, offering compact representations and robustness to background, appearance, and viewpoint~\cite{DS-STGCN,DeGCN}, making it well-suited for human-centered tasks.

Current STAS methods primarily model temporal and spatial relationships while supervising the predicted frame-level one-hot classification labels for both frame-wise accuracy and temporal smoothness. Specifically, in terms of modeling, existing approaches often utilize Graph Convolutional Networks (GCN)~\cite{StackGCN, MS-GCN, IDT-GCN, DeST, LaSA} or attention mechanisms~\cite{STGA-Net, SFA} to capture inter-joint relationships, and Temporal Convolution Networks (TCN)~\cite{StackGCN, MS-GCN, IDT-GCN, DeST} or attention mechanisms~\cite{STGA-Net, DeST, LaSA} to model long-range inter-frame dependencies. For supervision, classification predictions for each frame are guided by a categorical cross-entropy loss to minimize prediction errors, while a smoothing loss~\cite{MS-TCN, ASRF} penalizes over-segmentation errors by enforcing temporal consistency between consecutive frames.

Despite the success of current STAS methods, their lack of prior guidance limits the ability to capture the intrinsic correlations among actions as well as joints, resulting in a less nuanced understanding of movements compared to human-level perception. In fact, semantic differences and correlations exist both among various actions and joints. For instance, the relationships between the ``left hand," ``right hand," ``left elbow," and ``head" differ, and these inter-joint correlations subtly vary across actions. However, traditional GCN-based modeling methods struggle to effectively capture these correlations and adapt them to different actions.
Moreover, actions like ``walking" and ``running" are more semantically similar to each other than to ``sitting." Traditional one-hot encoding supervision, however, fails to capture these differences and relationships among actions, treating all categories as equally distinct.
Inspired by previous works~\cite{ActionCLIP, Bridge-prompt, GAP, LA-GCN}, this study proposes leveraging text-derived priors from large language models (LLMs) to enhance both modeling and supervision for improved semantic understanding of movements. Specifically, text-derived joint priors are utilized to establish fine-grained, action-adaptive inter-joint relationships, while text-derived action priors are employed to refine the semantic distributions and correlations among actions during supervision.

In addition, this study identifies two critical issues in data processing. First, the network exhibits an imbalanced perception of joints, failing to fully leverage the contributions of each joint. Second, although skeletal data is viewpoint-invariant, fixed skeleton orientations render the network sensitive to directional changes and unable to comprehend that actions are independent of orientation. However, joint occlusion and axial rotation help mitigate these limitations, thereby enhancing model robustness and generalization, and aligning more closely with human recognition intent.

This paper proposes the Text-Derived Relational Graph-Enhanced Network (TRG-Net) to enhance the understanding of intrinsic relationships in movements. TRG-Net consists of three key components: modeling, supervision, and data processing. For modeling, the Dynamic Spatio-Temporal Fusion Modeling (DSFM) method uses BERT~\cite{BERT}-encoded joint text embeddings to construct Text-Derived Joint Graphs (TJG), enabling fine-grained spatial modeling with dynamic channel- and frame-level adaptations. It also introduces spatio-temporal fusion to preserve core features during temporal modeling. 
For supervision, the Absolute-Relative Inter-Class Supervision (ARIS) method employs contrastive learning between BERT-encoded action text embeddings and output action features to supervise the feature distribution. It also constructs Text-Derived Action Graphs (TAG) from these embeddings to regularize the relative inter-class relationships.
For data processing, the Spatial-Aware Enhancement Processing (SAEP) method is a data augmentation technique that includes random joint occlusion and axial rotation, further enhancing model performance and generalization.

Validation on four public datasets—PKU-MMD (X-sub and X-view)~\cite{PKU-MMD}, LARa~\cite{LARA}, and MCFS-130~\cite{MCFS}—shows that TRG-Net achieves state-of-the-art results, significantly outperforming existing methods while maintaining reasonable efficiency. The contributions are as follows: 

\begin{itemize}
\item Modeling: We propose the Dynamic Spatio-Temporal Fusion Modeling  (DSFM) method, leveraging text-derived joint graphs, dynamic adaptations and spatio-temporal fusion for fine-grained modeling.
   
\item Supervision: We introduce the Absolute-Relative Inter-Class Supervision (ARIS) method, utilizing text-derived action graphs and action embeddings to enforce supervision on feature distribution and inter-class correlations.

\item Data Processing: We present the Spatial-Aware Enhancement Processing (SAEP) method, using random joint occlusion and axial rotation to enhance generalization. 
\end{itemize}


The remainder of this paper is organized as follows: Section~\ref{sec:rela_works} provides a review of previous works on action segmentation and vision-language multi-modal learning. Section~\ref{sec:methods} presents the proposed TRG-Net, followed by experiments of performance comparisons and effectiveness analyses in Section~\ref{sec:experiment}. Finally, Section~\ref{sec:Conclusion} concludes the paper.

\section{Related Works}
\label{sec:rela_works}

\subsection{Temporal Action Segmentation}

\begin{figure*}[t]
  \centering
   \includegraphics[width=0.9\linewidth]{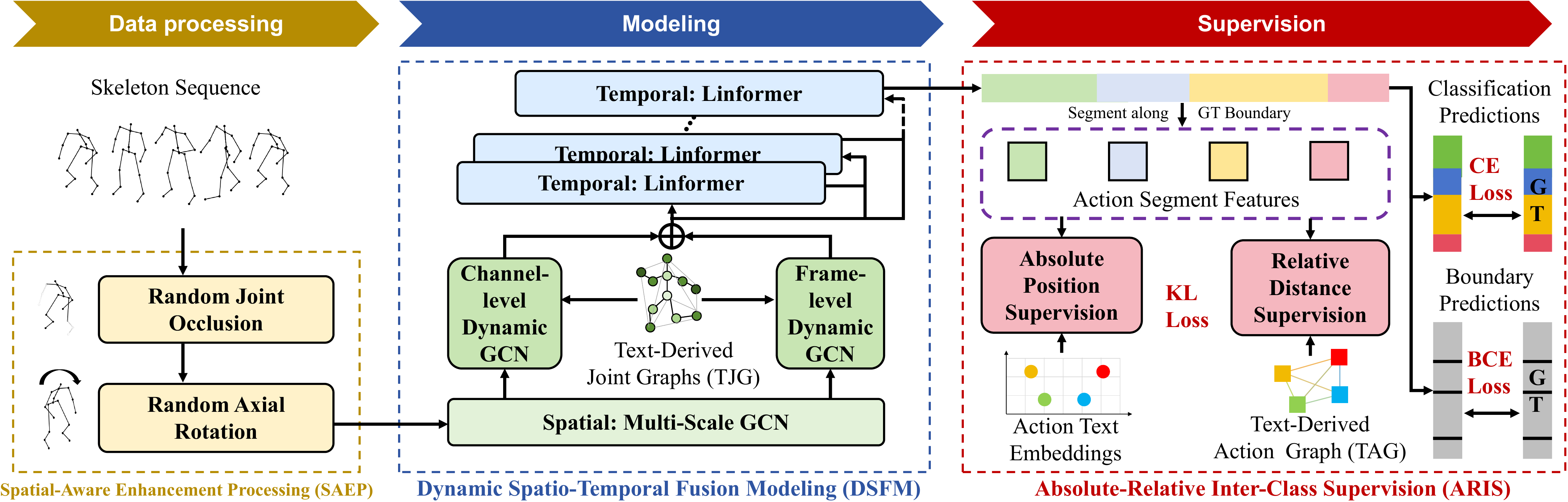}

   \caption{Overview of the TRG-Net. The TRG-Net consists of modeling, supervision, and data processing. The Dynamic Spatio-Temporal Fusion Modeling (DSFM) employs Text-Derived Joint Graphs (TJG) to support spatial dynamic modeling and achieves spatio-temporal fusion. The Absolute-Relative Inter-class Supervision (ARIS) utilizes Text-Derived Action Graphs (TAG) for relational supervision of action segments and action text embeddings for distribution supervision. Additionally, the Spatial-Aware Enhancement Processing (SAEP) method is introduced to further enhance generalization.}
   \label{fig:2}
\end{figure*}

\subsubsection{Video-based Temporal Action Segmentation}

VTAS methods take sequences of RGB or optical flow features extracted from video as input and utilize long-term temporal modeling to capture temporal relationships.  Temporal Convolutional Networks (TCN) are the mainstream in VTAS due to their effectiveness in capturing long-term temporal dependencies. Lea et al.~\cite{TCN} introduced the encoder-decoder TCN and the multi-layer dilated TCN, which was later refined by MS-TCN~\cite{MS-TCN} with a multi-stage refinement strategy. A series of improvements to TCN-based models followed, such as dual dilated layers~\cite{MS-TCN++}, the exploration of effective receptive field combinations~\cite{RF-Next}, the temporal series pyramid model with local burr suppression~\cite{ETSN}, coarse-to-fine multi-scale fusion strategies~\cite{C2F-TCN}, action boundary-aware methods~\cite{BCN,ASRF}, denoising diffusion models~\cite{Diffusion}, and the dilation passing and reconstruction method~\cite{DPRN}. Additionally, multimodal representation learnings~\cite{Aspnet} and prompt learning techniques~\cite{Bridge-prompt} are proposed. Graph-based temporal reasoning methods~\cite{GTRM,DTGRM,Semantic2Graph}, which treat frames or segments as graph nodes and apply graph convolution to extract semantic relations, have also achieved notable results. Meanwhile, other research explored the feasibility of Transformers in action segmentation. ASFormer~\cite{Asformer} was the first to introduce the Transformer to this field, and UVAST~\cite{UVAST} pioneered a sequence-to-sequence approach that outputs segment labels and frame counts rather than frame-level labels. Further Transformer-based architectures have been proposed, such as the U-shape pure Transformer~\cite{EUT}, and sparse attention to capture complete video context~\cite{LTContext}. Other approaches leveraging Transformer characteristics include boundary-aware query voting strategies~\cite{BaFormer} and frame-action cross-attention using action tokens~\cite{FACT}.

\subsubsection{Skeleton-based Temporal Action Segmentation}

STAS methods use sequences of human skeleton features as input, modeling spatial relationships among joints and temporal dependencies among frames separately. Mainstream approaches typically combine Graph Convolutional Networks (GCN) for spatial modeling with Temporal Convolutional Networks (TCN) for temporal modeling. Inspired by the ST-GCN~\cite{ST-GCN} in skeleton-based action recognition, Ghosh et al.~\cite{StackGCN} proposed a stacked hourglass architecture for action segmentation, while Filtjens et al.~\cite{MS-GCN} introduced the MS-GCN to refine predictions in multiple stages. A series of improvements followed, including the Involving Distinguished Temporal GCN approach~\cite{IDT-GCN}, multi-scale methods with motion-awareness and temporal enhancement~\cite{MTST-GCN}, and methods that decouple spatio-temporal interactions~\cite{DeST}. Other techniques focused on novel processing strategies, such as trajectory primitives and geometric features~\cite{Tai_Chi}, latent action composition~\cite{LAC}, and motion interpolation and action synthesis~\cite{CTC}. Several methods also explored attention mechanisms using Transformer models, such as fine-grained spatial focus attention~\cite{SFA} and spatio-temporal graph attention~\cite{STGA-Net}. Additionally, Ji et al.~\cite{LaSA} introduced LaSA, leveraging language modalities to assist in skeleton action understanding. However, existing methods fail to leverage the rich text relations and overlook spatial generalization. TRG-Net addresses these by using text-derived graph and spatial-aware methods to improve performance.

\subsection{Vision-language Multi-modal Representation Learning.}

Advances in Natural Language Processing (NLP), particularly with Large Language Models (LLM) such as GPT-3~\cite{GPT3} and BERT~\cite{BERT}, have significantly influenced the development of vision-language multi-modal representation learning in computer vision. Vision-language models (VLM) like CLIP~\cite{CLIP} and ALIGN~\cite{ALIGN} use contrastive learning to effectively align text and visual information, enhancing semantic understanding in downstream tasks. Several works have extended these models for few-shot learning, such as CoOp~\cite{CoOp}, which improves textual prompts by making them learnable, and CLIP-Adapter~\cite{Clip-adapter}, which adds learnable layers to both branches. For skeleton motion, MotionCLIP~\cite{MotionCLIP} aligns the human motion manifold with CLIP spaces. In video-based action recognition, ActionCLIP~\cite{ActionCLIP} is the first to introduce vision-language multi-modal learning to action recognition. In skeleton-based action recognition, GAP~\cite{GAP} supervises the skeleton encoder using textual descriptions of limb movements, while LA-GCN~\cite{LA-GCN} constructs prior graphs using joints and action texts to help learn the relations between them. Additionally, MMCL~\cite{MMCL} uses multi-modal LLM to enhance skeleton feature learning by incorporating both RGB and textual features. In video-based action segmentation, Bridge-Prompt~\cite{Bridge-prompt} uses contrastive learning between sequential action prompts and video segment features to train the video encoder, while UnLoc~\cite{Unloc} fuses class text information with frame features for modeling. In skeleton-based action segmentation, LaSA~\cite{LaSA} utilizes the language modality to assist in learning correlations and differences among joints and actions. Inspired by these methods, TRG-Net leverages text-derived relational graphs and embeddings generated by LLM to enhance both the modeling and supervision.

\section{Method}
\label{sec:methods}

STAS processes skeleton sequences denoted as \( X_{1:T} = [x_1, x_2, \dots, x_T] \in \mathbb{R}^{C_0 \times T \times V} \), where \( x_t \in \mathbb{R}^{C_0 \times V} \) represents the skeletal frame at time \( t \), with \( V \) and \( C_0 \) denoting the number of joints and channels, respectively. STAS maps the input \( X_{1:T} \) to the label space defined by the class set \( Q \) as \( Y_{1:T} = [y_1, y_2, \dots, y_T] \in \mathbb{R}^{Q \times T} \), where \( y_t = \{0,1\}^Q \) indicates the one-hot action label for the \( t \)-th frame.

This section outlines the overall design of TRG-Net, as depicted in Fig.~\ref{fig:2}. This work primarily leverage text-derived relational graphs to perform Dynamic Spatio-Temporal Fusion Modeling (DSFM) and Absolute-Relative Inter-Class Supervision (ARIS). Additionally, the novel Spatial-Aware Enhancement Processing (SAEP) method and the overall framework with the total loss is introduced.

\subsection{Text-derived Relational Graph}
\label{subsec3.1}

\begin{figure}[t]
  \centering
   \includegraphics[width=0.95\linewidth]{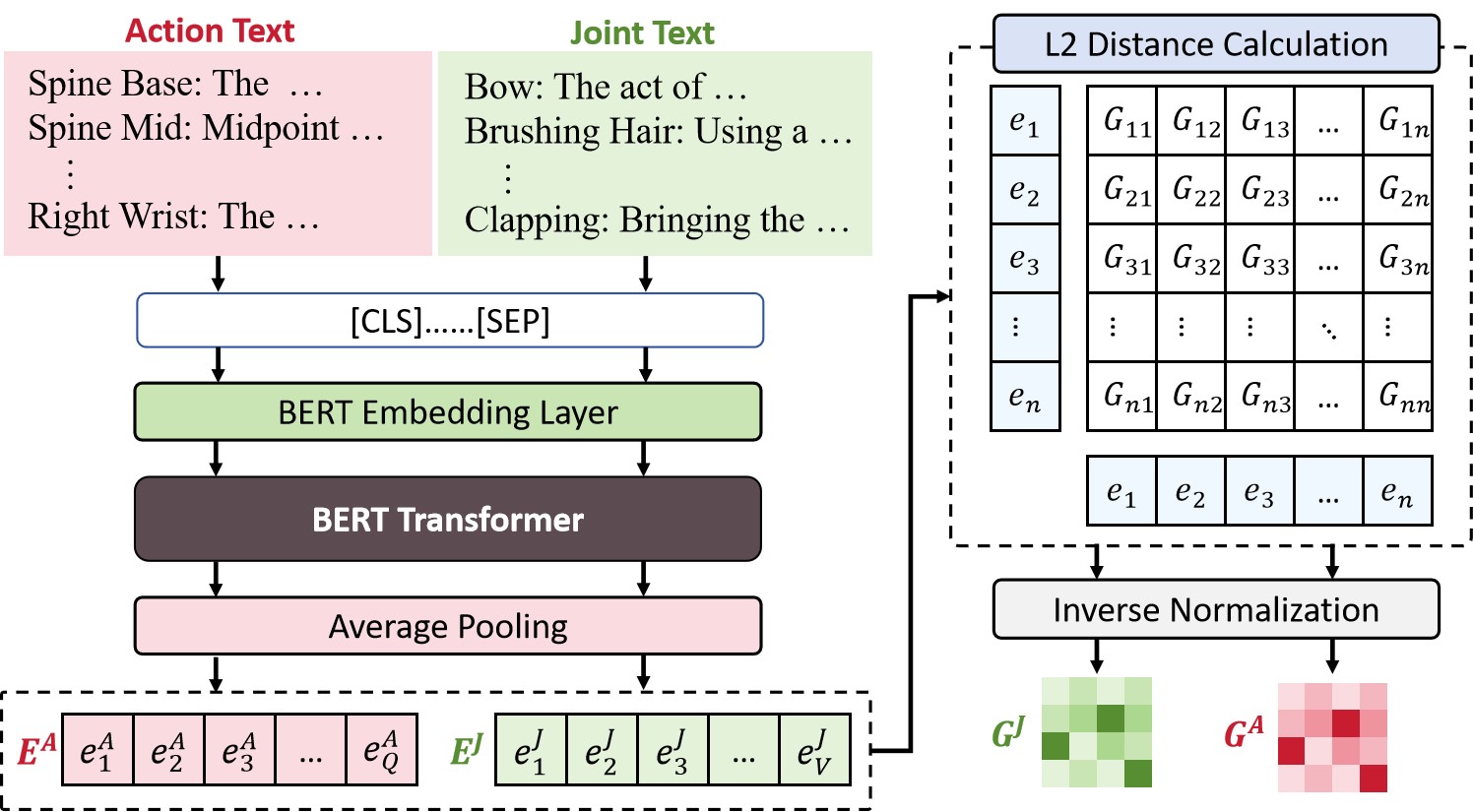}

   \caption{Generation of Text-Derived Relational Graphs. The descriptions of each joint and action are input into BERT to obtain the corresponding text embeddings. These embeddings are then processed using L2 distance calculations and inverse normalization to generate the TJG and TAG, which represent the semantic relationships between joints and between actions.}
   \label{fig:3}
\end{figure}

This study primarily employs the text-derived relational graphs to enhance modeling and supervision, which consists of Text-Derived Joint Graphs (TJG) and Text-Derived Action Graphs (TAG). These prior graphs are generated via BERT~\cite{BERT} (A large language model), utilizing joint and action class descriptions, as shown in Fig.~\ref{fig:3}. 
During the modeling and supervision processes, these fine-grained inter-class relational graphs serve as auxiliary inputs, providing intrinsic cues.

\textbf{Joint and Action Text Descriptions.}
Initially, we leverage the advanced comprehension capabilities of GPT-4~\cite{GPT4} to generate descriptive text for each joint and action. By prompting GPT-4 with the name of each joint or action, we obtain \( V \) joint descriptions and \( Q \) action descriptions, such as “Right Hip: The joint where the right thigh connects to the pelvis.”

\textbf{Joint and Action Text Embeddings.}
Each of the \( V \) joint descriptions and \( Q \) action descriptions is prefixed with the [CLS] token and suffixed with the [SEP] token. Subsequently, each word token in the sentences is converted to an index based on the vocabulary, producing \( I^J \in \mathbb{R}^{V \times H} \) and \( I^A \in \mathbb{R}^{Q \times H} \), where \( H \) denotes the maximum sentence length, and shorter sentences are zero-padded. These matrices are then input into pre-trained embedding layer and Transformers of BERT to obtain the modeled embeddings \( E^J_1 \in \mathbb{R}^{V \times H \times Ct} \) and \( E^A_1 \in \mathbb{R}^{Q \times H \times Ct} \). The word embeddings in each sentence are then averaged through pooling, resulting in Joint/Action Text Embeddings \( E^J \in \mathbb{R}^{V \times Ct} \) and \( E^A \in \mathbb{R}^{Q \times Ct} \).

\textbf{Text-Derived Relational Graphs.}
The Text-Derived Relational Graphs are computed based on the internal embeddings of \( E^J \) and \( E^A \). For the Text-Derived Joint Graphs (TJG), the pairwise Euclidean (L2) distances between the \( V \) joint embeddings \( e^J_i \in \mathbb{R}^{Ct} \) within \( E^J = [e^J_1, e^J_2, \dots, e^J_V] \) are calculated, resulting in the graph \(\widetilde{G}^J \in \mathbb{R}^{V \times V} \):

\begin{equation}
\widetilde{G}^J_{ij} = \|e^J_i - e^J_j\|_2 = \sqrt{\sum_{k=1}^{n} (E^J_{ik} - E^J_{jk})^2},
\end{equation}
where \( \widetilde{G}^J_{ij} \) is the element at the \( i \)-th row and \( j \)-th column of \( \widetilde{G}^J \), \( e^J_i \) denotes the \( i \)-th row vector of \( E^J \), and \( E^J_{ik} \) is the element in the \( i \)-th row and \( k \)-th column of \( E^J \). The L2 norm is denoted by \( \|\cdot\|_2 \). 
Subsequently, the matrix \( \widetilde{G}^J \) undergoes inverse normalization to form TJG, denoted as \( G^{J} \):
\begin{equation}
G^{J} = 1 - \frac{\widetilde{G}^J - \widetilde{G}^J_{\text{min}}}{\widetilde{G}^J_{\text{max}} - \widetilde{G}^J_{\text{min}}},
\end{equation}
where \( \widetilde{G}^J_{\text{min}} \) and \( \widetilde{G}^J_{\text{max}} \) represent the minimum and maximum values of \( \widetilde{G}^J \), respectively. Therefore, the closer the L1 distance between two embeddings, the closer the corresponding value in TJG approaches 1, while larger distances approach 0. Similarly, the Text-Derived Action Graph (TAG) is derived from \( E^A \) via L2 distance calculation and inverse normalization, resulting in \( G^{A} \in \mathbb{R}^{Q \times Q} \).

\subsection{Dynamic Spatio-Temporal Fusion Modeling}
\label{subsec3.2}

\begin{figure}[t]
  \centering
   \includegraphics[width=0.95\linewidth]{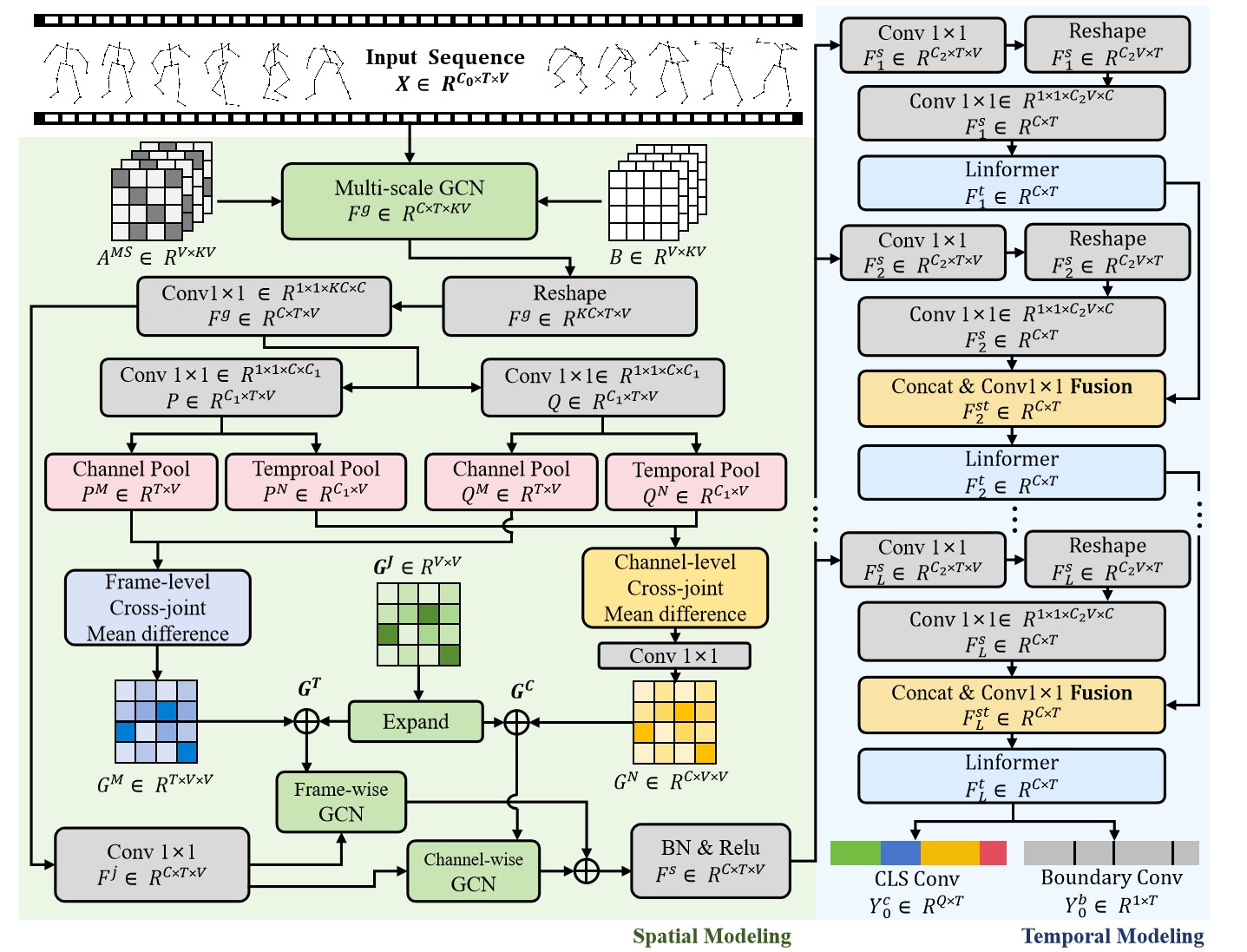}

   \caption{Dynamic Spatio-Temporal Fusion Modeling. The process begins with multi-scale GCN for initial spatial modeling. Then the combination of TJG and feature-derived channel- and frame-level dynamic joint graphs is used for fine-grained adaptive spatial modeling. Subsequently, adaptive weight allocation through convolution merges joint features with channel features, while linear transformers conducts multi-layer temporal modeling to establish inter-frame relationships. Spatial features with weighted adjustments are fused after each temporal layer, progressively integrating core spatio-temporal features.}
   \label{fig:4}
\end{figure}

This section introduces the modeling of TRG-Net, termed Dynamic Spatio-Temporal Fusion Modeling (DSFM), as shown in Fig.~\ref{fig:4}. In spatial modeling,  DSFM leverages Text-Derived Joint Graphs (TJG) to provide semantic-level guidance for the spatial feature, integrating feature-derived dynamic graphs at channel and frame levels for fine-grained adaptive spatial modeling. In temporal modeling, a multi-level spatio-temporal fusion approach progressively combines core spatial features extracted from various perspectives with temporal modeling  features from each layer, facilitating the effective retention and combination of essential spatio-temporal characteristics.

\textbf{Text-based adaptive spatial modeling.}
TRG-Net first adopts a multi-scale GCN inspired by MS-G3D~\cite{MS-G3D} and DeST~\cite{DeST} to capture spatial dependencies among joints, generating \(F^{g}\). Then, the text-based adaptive spatial modeling utilizes TJG and integrates channel- and frame-level adaptive graphs to enable fine-grained spatial modeling.
Specifically, the model first processes \(F^{g} \in \mathbb{R}^{C \times T \times V}\) through two convolutional heads to obtain \(P, Q \in \mathbb{R}^{C_1 \times T \times V}\). Then, \(P\) and \(Q\) undergo  channel pooling to produce \(P^{M}, Q^{M} \in \mathbb{R}^{T \times V}\), and temporal pooling to obtain \(P^{N}, Q^{N} \in \mathbb{R}^{C_1 \times V}\). Subsequently, frame-level adaptive graph \(G^{M} \in \mathbb{R}^{T \times V \times V}\) and channel-level adaptive graph \(G^{N} \in \mathbb{R}^{C_1 \times V \times V}\) are computed using cross-joint mean differences:
\begin{equation}
G^{M}_{t,i,j} = P^{M}_{t,i} - Q^{M}_{t,j}, \quad
G^{N}_{c,i,j} = P^{N}_{c,i} - Q^{N}_{c,j},
\end{equation}
where \(G^{M}_{t,i,j}\) represents the element at position \(i,j\) of \(t\)-th frame in \(G^{M}\), and \(F^{M}_{t,i}\) denotes the \(i\)-th element of \(t\)-th frame in \(F^{M}\); the other terms are defined similarly. The TJG \(G^{J} \in \mathbb{R}^{V \times V}\) is then expanded along both frame and channel dimensions and added to \(G^{M}\) and \(G^{N}\), yielding the Frame-adaptive Text-derived Joint Graph (FTJG) \(G^{T} \in \mathbb{R}^{T \times V \times V}\) and the Channel-adaptive Text-derived Joint Graph (CTJG) \(G^{C} \in \mathbb{R}^{C_1 \times V \times V}\).
Subsequently, \(F^j \in \mathbb{R}^{C \times T \times V}\), derived from \(F^{g}\) via convolution, undergoes graph convolution through both \(G^{T}\) and \(G^{C}\), leading to the final fusion feature:
\begin{equation}
F^{s} = \mathrm{ReLU}\left(\mathrm{BN}\left[F^j \cdot G^{T} + F^j \cdot G^{C}\right]\right),
\end{equation}
where \(\cdot\) denotes matrix multiplication along the joint dimension \(V\). Finally, batch normalization and ReLU activation yield the spatially modeled feature \(F^{s} \in \mathbb{R}^{C \times T \times V}\).

\textbf{Temporal Modeling with Spatio-temporal Fusion.}
First, the feature \(F^{s}\) undergoes convolution to produce \(F^{s}_1 \in \mathbb{R}^{C_2 \times T \times V}\), followed by a reshape and convolution to merge the spatial dimension \(V\) into the channel dimension \(C_2\), resulting in \(F^{s}_1 \in \mathbb{R}^{C \times T}\). The convolution is adaptively optimized for each channel, providing distinct weights for different spatial dimensions \(V\), thereby retaining the core spatial features in an adaptive manner. This transformed feature is then input into a linear transformer~\cite{Linformer1} for temporal modeling, producing \(F^{t}_1\).

Next, the same temporal layer is repeated for \(L-1\) times to progressively enhance inter-frame dependencies. Before each subsequent temporal modeling step, the temporal features \(F^{t}_{l-1}\) from the previous layer are fused with the spatial features \(F^{s}_l\). Specifically, the initial spatial feature \(F^{s}\) is passed through different convolutional heads, reshaped, and convolved to merge the spatial dimension \(V\) into the channel dimension \(C_2\), producing \(F^{s}_l \in \mathbb{R}^{C \times T}\). Due to the differences in convolutional heads, each layer generates distinct spatial fusion weights, allowing each \(F^{s}_l\) to represent spatial features with varying levels of integration, emphasizing different core spatial characteristics. The spatial feature \(F^{s}_l\) and the temporal feature \(F^{t}_{l-1}\) are combined to produce \(F^{st}_l \in \mathbb{R}^{C \times T}\), defined as:
\begin{equation}
F^{st}_l = \mathrm{GeLU}\left[\left(F^{s}_l \oplus F^{t}_{l-1}\right) \cdot W_f \cdot W_l\right] + F^{t}_{l-1},
\end{equation}
where \(\oplus\) denotes channel concatenation, \(W_f \in \mathbb{R}^{1 \times 1 \times 2C \times C}\) is a convolution operator, and \(W_l \in \mathbb{R}^{C \times C}\) is a linear layer. This fusion mechanism adaptively fine-tunes the contribution of both spatial and temporal features. The fused output \(F^{st}_l\) is then input into the next linear transformer for further temporal modeling, resulting in \(F^{t}_l\).

After \(L-1\) spatio-temporal fusion operations and \(L\) temporal layers, the final feature \(F^{t}_L \in \mathbb{R}^{C \times T}\) is fed into both a classification head and a boundary regression head, yielding frame-level class predictions \(Y^{c}_0 \in \mathbb{R}^{Q \times T}\) and frame-level boundary predictions \(Y^{b}_0 \in \mathbb{R}^{1 \times T}\).

\subsection{Absolute-Relative Inter-Class Supervision}
\label{subsec3.3}

\begin{figure}[t]
  \centering
   \includegraphics[width=0.95\linewidth]{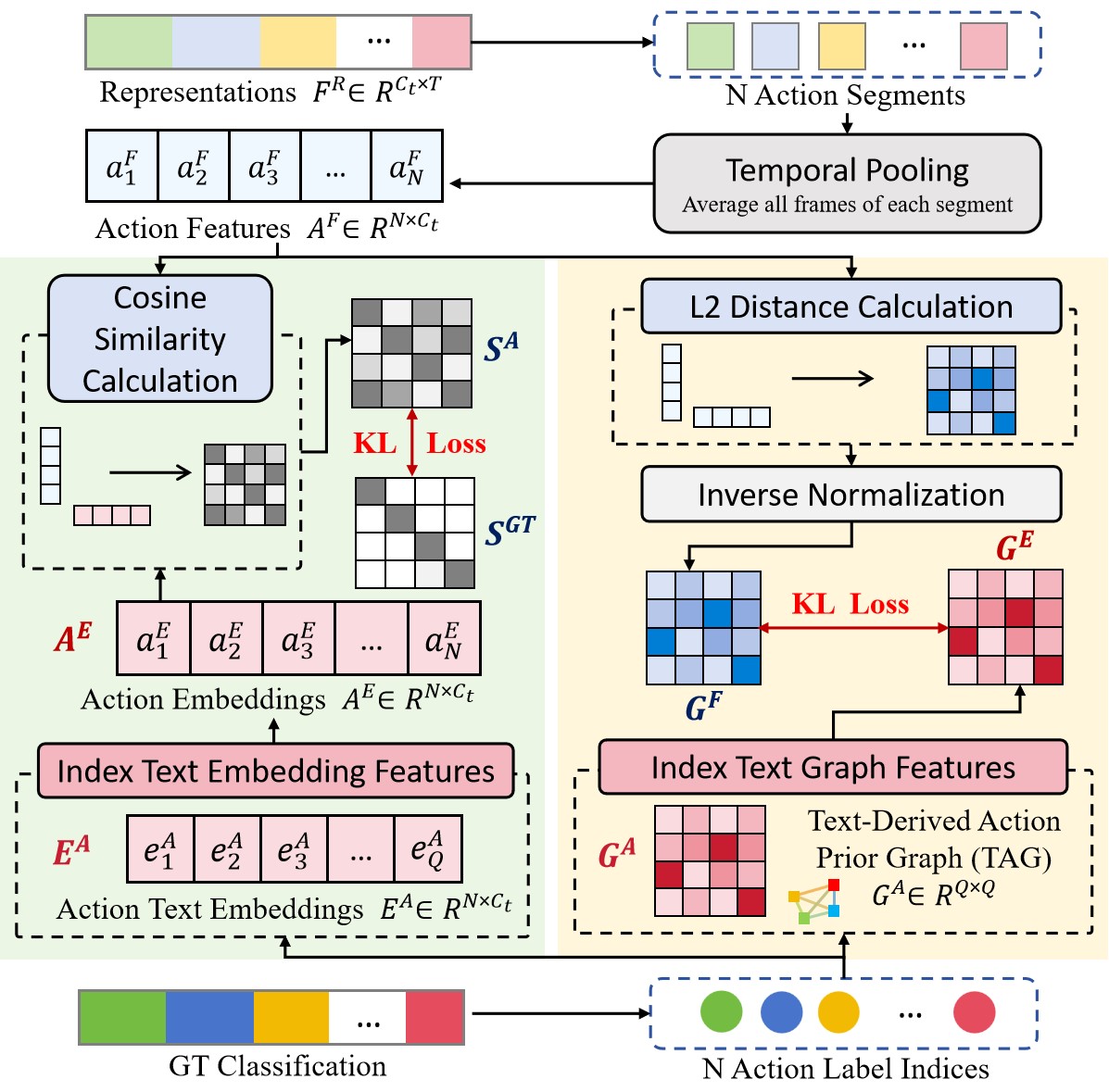}

   \caption{Absolute-Relative Inter-Class Supervision. First, representations are segmented based on ground truth boundaries and pooled to obtain individual action features. Absolute supervision (left) compares each action feature with its corresponding text embedding via cosine similarity, optimizing class distributions through contrastive learning. Relative supervision (right) aligns computed action feature relations with the semantic relations in TAG, refining inter-class relative relationships.}
   \label{fig:5}
\end{figure}

This section presents the supervision of TRG-Net, referred to as Absolute-Relative Inter-Class Supervision (ARIS). ARIS innovatively employs action text embeddings and the subsequently generated Text-Derived Action Graphs (TAG) to facilitate absolute-relative semantic-level supervision of the output representations, as shown in Fig.~\ref{fig:5}. This approach optimizes class distribution while emphasizing the relationships among action classes. Specifically, absolute supervision utilizes contrastive learning on action-text pairs, formed by action features and their corresponding action text embeddings, to enhance action feature distributions. In contrast, relative supervision computes the relations among action features and aligns them with the TAG to refine the inter-class relationships.

\textbf{Absolute Distribution Supervision.} Absolute distribution supervision utilizes an approach akin to CLIP~\cite{CLIP}, , where each action segment feature undergoes contrastive learning with its corresponding text embedding. This facilitates the alignment of skeleton features with semantically well-distributed text features,  improving the action feature distributions, clustering and inter-class discrimination. 

Specifically, the final output features \(F_L^t \in \mathbb{R}^{C \times T}\) are projected to a high-dimensional latent space, resulting in feature representations \(F^R \in \mathbb{R}^{C_t \times T}\).  \(F^R\) are segmented according to the ground truth boundaries into \(N\) action segment features, each globally averaged along temporal dimension to produce a set of  \(N\) action features \(A^F \in \mathbb{R}^{C_t \times N}\). Similarly, based on the class labels of the \(N\) action features, the corresponding embeddings from \(Q\) classes of action text embeddings \(E^A \in \mathbb{R}^{C_t \times Q}\) are indexed, producing a set of \(N\) action embeddings \(A^E \in \mathbb{R}^{C_t \times N}\). At this stage, a contrastive learning is applied between action features \(A^F\) and action embeddings \(A^E\) to align \(A^F\) with \(A^E\), thereby learning the semantic distribution among action classes at the textual level. 

Then, the contrastive learning computes the cosine similarity \(sim(\cdot)\) between each embedding in \(A^F\) and each embedding in \(A^E\), yielding a similarity matrix \(S^A \in \mathbb{R}^{N \times N}\):
\begin{equation}
S^A = 
\begin{bmatrix} 
    sim(A^F_1,A^E_1)  & \cdots & sim(A^F_1,A^E_N)\\
    \vdots & \ddots & \vdots\\
    sim(A^F_N,A^E_1)  & \cdots & sim(A^F_N,A^E_N)
\end{bmatrix},
\end{equation}
where cosine similarity \(sim(z_i,z_j) = (z_i \cdot z_j) / (|z_i| |z_j|)\). Softmax normalization is then applied to \(S^A\) along both rows and columns, yielding \(S^A_f \in \mathbb{R}^{N \times N}\) normalized over visual feature dimensions and \(S^A_e \in \mathbb{R}^{N \times N}\) normalized over text embedding dimensions. The ground truth similarity matrix \(S^{GT} \in \mathbb{R}^{N \times N}\) is defined such that elements corresponding to the same action class (positive pairs) have a value of 1, while negative pairs have a value of 0. The optimization is achieved by minimizing the Kullback-Leibler (KL) divergence between the similarity matrix and the ground truth:
\begin{equation}
\mathcal{D}_{KL}(U\| W) = \frac{1}{N^2} \sum_{i=1}^{N} \sum_{j=1}^{N} U_{ij} \log \left( \frac{U_{ij}}{W_{ij}} \right),
\end{equation}
where \( U, W \in \mathbb{R}^{N \times N} \). This contrastive loss concurrently optimizes the two similarity matrices \(S^A_f\) and \(S^A_e\):
\begin{equation}
\mathcal{L}_{abs} = \frac{1}{2}[\mathcal{D}_{KL}(S^A_f\| S^{GT})+\mathcal{D}_{KL}(S^A_e\| S^{GT})].
\end{equation}

\textbf{Relative Distance Supervision}
While absolute distribution supervision effectively learns a desirable distribution, its rigid alignment of features can be excessively constraining, relying heavily on high model parameters and extensive training data, which may lead to fitting difficulties in typical models. Relative distance supervision focuses on optimizing the relative relations among action classes rather than the specific distribution positions of each class, aligning more closely with human intuitive understanding of actions and highlighting similarities and differences among diverse actions. This supervision enhances relative relationships while mitigating unnecessary redundant optimization weights from absolute supervision.

Specifically, relative supervision also utilizes action features \(A^F \in \mathbb{R}^{C_t \times N}\), derived from \(F^R\) through segmentation and average pooling. By calculating the L2 distances among the \(N\) action features in \(A^F\) and applying inverse normalization, we obtain a feature relational graph \(G^F \in \mathbb{R}^{N \times N}\), where greater L2 distances between two action features yield values approaching 0 in the graph, while closer distances approach 1. Simultaneously, based on the class labels of the \(N\) action features, corresponding text-derived relational elements are indexed from TRG \( G^{A} \in \mathbb{R}^{Q \times Q}\), resulting in a text-level guidance graph \(G^E \in \mathbb{R}^{N \times N}\). The alignment optimization of the feature relational graph \(G^F\) with the text-level guidance graph \(G^E\) is achieved by minimizing the KL divergence:
\begin{equation}
\mathcal{L}_{rel} = \mathcal{D}_{KL}(G^F\| G^E).
\end{equation}

\subsection{Spatial-Aware Enhancement Processing}
\label{subsec3.4}

In addition to text-based modeling and supervision, TRG-Net introduces a spatial-aware data augmentation strategy aimed at improving the spatial generalization. This study identifies a gap in the skeleton-based action segmentation field, where data augmentation techniques common in image processing have been underexplored. However, such techniques can mitigate overfitting and improve generalization, robustness, and performance. Inspired by human recognition of actions is unaffected by orientation variation or body part occlusion, we propose random joint occlusion and axial rotation as augmentation methods. 

\textbf{Random Joint Occlusion.} 
This study posits that models should recognize actions based solely on the available joints, even with random occlusion. Training under such conditions can enhance the contribution of each joint to model understanding, thereby elevating the importance of previously underappreciated joints. 
Specifically, during training, each input instance \(X \in \mathbb{R}^{C_0 \times T \times V}\) undergoes random occlusion of 0-50\% of fixed joints, generating a mask \(M^v \in \mathbb{R}^{V}\) where 0-50\% of elements are randomly set to 0 and the remainder to 1. \(M^v\) is then expanded to the dimensions \(C_0, T\), resulting in \(M^X \in \mathbb{R}^{C_0 \times T \times V}\), which is used for random occlusion:  
\begin{equation}
X^{occ} = X \odot M^X,
\end{equation}
where \(\odot\) denotes element-wise multiplication, and \(X^{occ}\) is the occluded skeleton sequence. Note that the occluded joints are the same across all frames in a sequence, but vary across sequences. 

\textbf{Random Axial Rotation.}
Human recognition of actions is independent of the orientation, demonstrating the ``isotropic" nature of actions. In real-world scenarios, actions remain identifiable regardless of facing direction.  Thus, models should have a generalized perception that is unaffected by direction.  To achieve this, each input instance \(X \in \mathbb{R}^{C_0 \times T \times V}\) is rotated around the axial direction by a random angle \(\theta \in (0,2\pi)\) using a rotation matrix  \(\mathcal{R}(\theta) \in \mathbb{R}^{3 \times 3}\). Since the channel dimension \(C_0\) is often a multiple of 3, encompassing combinations of 3D position, 3D displacement, 3D acceleration, and 3D posture, the transformation is applied to each set of three axes:  
\begin{equation}
X^{rot} =   
\begin{bmatrix} 
    \mathcal{R}(\theta)  & \cdots & 0\\
    \vdots & \ddots & \vdots\\
    0  & \cdots & \mathcal{R}(\theta)
\end{bmatrix}
\cdot X,
\end{equation}
where the left matrix has dimensions \(\mathbb{R}^{C_0 \times C_0}\), formed by multiple rotation matrices \(\mathcal{R}(\theta)\) along the diagonal, thereby facilitating the rotation of every three axes in \(X\). 

During training, a proportion \(\alpha\) of input sequences undergoes random joint occlusion, while another proportion \(\beta\) is subjected to axial rotation, with the remaining sequences left unaltered. During inference, both are not applied to ensure consistency and accuracy.

\subsection{Overall Framework}
\label{subsec3.5}

\textbf{Overall Modeling.}
In addition to DSFM, the complete model further refines the class predictions \(Y^{c}_0 \in \mathbb{R}^{Q \times T}\) and boundary predictions \(Y^{b}_0 \in \mathbb{R}^{1 \times T}\) generated by DSFM, following the approach in ~\cite{ASRF,DeST,LaSA}. For \(Y^{c}_0\), the class predictions are passed through a class prediction branch with \(S^c\) stages, where each stage contains 10 layers of linear transformers for cross-attention, resulting in final class predictions \(Y^{c}_{F} \in \mathbb{R}^{Q \times T}\). For \(Y^{b}_0\), the boundary predictions are refined via a boundary regression branch with \(S^{b}\) stages, each consisting of 10 layers of Dilated TCN for temporal modeling, yielding final boundary predictions \(Y^{b}_{F} \in \mathbb{R}^{1 \times T}\).

\textbf{Overall Supervision.}
In addition to the absolute supervision with \(\mathcal{L}_{abs}\) and relative supervision with \(\mathcal{L}_{rel}\), classic action segmentation supervision methods~\cite{ASRF} are also applied. These include action segmentation loss \(\mathcal{L}_{as}\) for class predictions of all stages, and boundary regression loss \(\mathcal{L}_{br}\) for boundary predictions of all stages. The action segmentation loss \(\mathcal{L}_{as}\)  for each stage comprises the frame-level classification cross-entropy loss \(\mathcal{L}_{ce}\) and the Gaussian Similarity-weighted Truncated Mean Squared Error (GS-TMSE) loss \(\mathcal{L}_{gs-tmse}\):
\begin{equation}
\begin{gathered}
\mathcal{L}_{as}= \mathcal{L}_{ce} + \mathcal{L}_{gs-tmse} 
= -\frac{1}{T} \sum_{t}{\log(\hat{y}_{t, \hat{c}})} \\
+ \frac{1}{TC} \sum_{t,c} e^{-\frac{\|\hat{y}_{t}-\hat{y}_{t-1}\|^2}{2\sigma^2}} {|log(\frac{\hat{y}_{t,c}}{\hat{y}_{t-1,c}})|^2},
\end{gathered}
\end{equation}
where \(\hat{y}_{t}\) represents the prediction for \(t\)-th frame, and \(\hat{y}_{t,\hat{c}}\) is the predicted probability for the ground truth class \(\hat{c}\) at time \(t\). The cross-entropy loss \(\mathcal{L}_{ce}\) supervises the per-frame classification, while \(\mathcal{L}_{gs-tmse}\) smooths adjacent frames to mitigate over-segmentation, with a maximum log difference threshold \(\tau = 4\) to prevent over-smoothing. The boundary regression loss \(\mathcal{L}_{br}\) for each stage is defined as the frame-level binary cross-entropy loss:
\begin{equation}
\mathcal{L}_{br} = -\frac{1}{T} \sum_{t} \left( b_t \log(\hat{b}_t) + (1 - b_t) \log(1 - \hat{b}_t) \right),
\end{equation}
where \(b_t\) is the ground truth boundary label (1 for boundary frames, 0 for others), and \(\hat{b}_t\) is the predicted boundary probability for \(t\)-th frame.

\section{Experiment}
\label{sec:experiment}

\subsection{Datasets}
The model undergoes performance evaluation and comparison across four publicly available datasets: PKU-MMD (X-sub and X-view)~\cite{PKU-MMD}, MCFS-130~\cite{MCFS}, and LARa~\cite{LARA}. 

\begin{itemize}
\item\textbf{PKU-MMD}~\cite{PKU-MMD} is a large-scale action dataset recorded at 30 Hz using Kinect V2 sensors, encompassing 3-axis position data for 25 joints. It contains 1,009 videos across 52 action categories, totaling 50 hours. This study utilizes two partition settings: X-sub, where the training and test data are from different videos, and X-view, where the training and test data are from different views. 

\item\textbf{LARa}~\cite{LARA} is a dataset focused on typical warehouse activities, recorded at 200 Hz using optical marker-based motion capture devices. It includes 3-axis position and orientation data for 19 joints, comprising 377 videos across 8 action categories, totaling 758 minutes. In this study, the 200 Hz recordings are downsampled to 50 Hz.

\item\textbf{MCFS-130}~\cite{MCFS} is a figure skating dataset captured at 30 Hz using OpenPose toolbox, which includes 2-axis position data for 25 joints. It consists of 271 videos categorized into 130 action classes, totaling 17.3 hours.
\end{itemize}

\subsection{Evaluation Metrics}
This study employs three classical evaluation metrics for action segmentation: Frame-wise Accuracy (Acc), Segmental Edit Scores, and Segmental F1 Scores at overlap thresholds of 0.10, 0.25, and 0.5, denoted as F1@\{10, 25, 50\}.

\begin{itemize}
\item\textbf{Frame-wise Accuracy (Acc)} measures the proportion of correctly predicted frames, providing a direct and sensitive assessment of performance. However, it tends to overlook over-segmentation errors.

\item\textbf{Segmental Edit Scores} quantify the minimum number of edits (insertions, deletions, and substitutions) required to transform one segment sequence into another. This metric evaluates the segment-level discrepancies between predictions and ground truth, effectively penalizing over-segmentation while being insensitive to segment length.

\item\textbf{Segmental F1 Scores} are calculated by comparing the Intersection-over-Union (IoU) of each predicted segment with its corresponding ground truth based on specified overlap thresholds. This approach computes precision and recall to get F1 scores for all predicted segments, thereby penalizing over-segmentation errors while maintaining insensitivity to minor temporal deviations.
\end{itemize}

\subsection{Implementation Details}
Regarding the channel of the model, the typical feature channel \(C = 64\). During the establishment of the adaptive graph, the channel \(C_1 = 16\). Before the merge of the dimension \(V\), the feature channel \(C_2 = 8\), while the the text embedding channel \(C_t = 768\). Additionally, the number of temporal modeling layers in DSTM is \(L = 10\). The subsequent class prediction branch consists of 1 stage, while the boundary regression branch comprises 2 stages. 
For the LARa, a learning rate of 0.001 and a batch size of 8 were used for training over 120 epochs. For the MCFS, a learning rate of 0.0005 and a batch size of 1 were employed for 300 epochs. For the PKU-MMD, a learning rate of 0.001 and a batch size of 8 were utilized for 300 epochs. All training and testing experiments were conducted on a single NVIDIA RTX 3090 GPU.

\subsection{Performance Comparison}

\subsubsection{Quantitative Comparison}

\begin{table*}[t]
    \footnotesize
    \centering
    \caption{Comparison with the latest results on PKU-MMD (X-sub) and PKU-MMD (X-view).  \textbf{Bold} and \underline{underline} indicate the best and second-best results in each column. FLOPs denote the computational complexity of the model when the input is $X \in \mathbb{R}^{12 \times 6000 \times 25 }$ on PKU-MMD dataset. }
    \begin{tabular}{c|l|cc|ccccc|ccccc}
        \toprule
        & Dataset & & & \multicolumn{5}{c|}{PKU-MMD (X-sub)} & \multicolumn{5}{c}{PKU-MMD (X-view)} \\
        \cmidrule(lr){5-9} \cmidrule(lr){10-14}
        & Metric & FLOPs$\downarrow$ & Param.$\downarrow$ & Acc$\uparrow$ & Edit$\uparrow$ & \multicolumn{3}{c|}{F1@\{10,25,50\}$\uparrow$} & Acc$\uparrow$ & Edit$\uparrow$ & \multicolumn{3}{c}{F1@\{10,25,50\}$\uparrow$} \\
        \midrule
        \multirow{7}{*}{\begin{sideways}\textbf{VTAS}\end{sideways}}
        & MS-TCN~\cite{MS-TCN} & 3.27G & 0.54M & 65.5 & - & - & - & 46.3 & 58.2 & 56.6 & 58.6 & 53.6 & 39.4 \\
        & MS-TCN++~\cite{MS-TCN++} & 5.37G & 0.90M & 66.0 & 66.7 & 69.6 & 65.1 & 51.5 & 59.1 & 59.4 & 61.6 & 56.1 & 42.0  \\
        & ETSN~\cite{ETSN} & 4.44G & 0.75M  & 68.4 & 67.1 & 70.4 & 65.5 & 52.0 & 60.7 & 57.6 & 62.4 & 57.9 & 44.3 \\
        & ASFormer~\cite{Asformer} & 6.34G & 1.04M & 68.3 & 68.1 & 71.9 & 68.2 & 54.5 & 64.8 & 62.6 & 65.1 & 60.0 & 45.8 \\
        & ASRF~\cite{ASRF} & 7.15G & 1.20M & 67.7 & 67.1 & 72.1 & 68.3 & 56.8 & 60.4 & 59.3 & 62.5 & 58.0 & 46.1 \\
        & C2F-TCN~\cite{C2F-TCN} & 11.4G & 4.54M & 68.2 & 58.4 & 63.0 & 58.6 & 46.7 & 64.9 & 54.3 & 57.3 & 51.7 & 40.4 \\
        & LTContext~\cite{LTContext} & 3.71G & 0.56M & 68.8 & 68.2 & 72.1& 68.5 & 55.4 & 62.3 & 59.6 & 62.9 & 57.8 & 42.6 \\
        
        \midrule
        \multirow{4}{*}{\begin{sideways}\textbf{STAS}\end{sideways}}
        & MS-GCN~\cite{MS-GCN} & 40.2G & 0.65M & 68.5 & - & - & - & 51.6 & 65.3 & 58.1 & 61.3 & 56.7 & 44.1 \\
        & CTC~\cite{CTC} & - & - & 69.2 & - & 69.9 & 66.4 & 53.8 & - & - & - & - & - \\
        & MTST-GCN~\cite{MTST-GCN} & 141.9G & 2.90M & 70.0 & 65.8 & 68.5 & 63.9 & 50.1 & 66.5 & 64.0 & 67.1 & 62.4 & 49.9 \\
        & DeST-{\tiny TCN}~\cite{DeST} & 6.19G & 0.78M & 67.6 & 66.3 & 71.7 & 68.0 & 55.5 & 62.4 & 58.2 & 63.2 & 59.2 & 47.6 \\
        & DeST-{\tiny Former}~\cite{DeST} & 8.17G & 1.11M & 70.3 & 69.3 & 74.5 & 71.0 & 58.7 & 67.3 & 64.7 & 69.3 & 65.6 & 52.0 \\
        & LaSA~\cite{LaSA} & 11.65G & 1.60M & \underline{73.5} & \underline{73.4} & \underline{78.3} & \underline{74.8} & \underline{63.6} & \underline{69.5} & \underline{67.8} & \underline{72.9} & \underline{69.2} & \underline{57.0} \\
        
        \cmidrule(lr){2-14}
        & \textbf{TRG-Net} & 11.0G & 1.36M & \textbf{75.6} & \textbf{73.9} & \textbf{79.1} & \textbf{76.4} & \textbf{65.8} & \textbf{75.1} & \textbf{73.1} & \textbf{78.9} & \textbf{76.3} & \textbf{65.1} \\
        \bottomrule
    \end{tabular}
    \label{tab:SOTA1}
\end{table*}

\begin{table*}[t]
    \footnotesize
    \centering
    \caption{Comparison with the latest results on LARa, MCFS-130. \textbf{Bold} and \underline{underline} indicate the best and second-best results in each column.}
    \begin{tabular}{c|l|ccccc|ccccc}
        \toprule
        & Dataset & \multicolumn{5}{c|}{LARa} & \multicolumn{5}{c}{MCFS-130} \\
        \cmidrule(lr){3-7} \cmidrule(lr){8-12}
        & Metric & Acc$\uparrow$ & Edit$\uparrow$ & \multicolumn{3}{c|}{F1@\{10, 25, 50\}$\uparrow$} & Acc$\uparrow$ & Edit$\uparrow$ & \multicolumn{3}{c}{F1@\{10, 25, 50\}$\uparrow$} \\
        \midrule
        \multirow{7}{*}{\begin{sideways}\textbf{VTAS}\end{sideways}}
        & MS-TCN~\cite{MS-TCN} & 65.8 & - & - & - & 39.6 & 65.7& 54.5& 56.4 & 52.2 & 42.5 \\
        & MS-TCN++~\cite{MS-TCN++} & 71.7 & 58.6 & 60.1 & 58.6 & 47.0 & 65.5 & 59.8 & 60.1 & 55.7 & 46.1  \\
        & ETSN~\cite{ETSN} & 71.9 & 58.4 & 64.3 & 60.7 & 48.1 & 64.6& 64.6 & 64.5 & 61.0 & 52.3 \\
        & ASFormer~\cite{Asformer} & 72.2 & 62.2 & 66.1 & 61.9 & 49.2 & 67.5& 69.1 & 68.3 & 64.0 & 55.1 \\
        & ASRF~\cite{ASRF} & 71.9 & 63.0 & 68.3 & 65.3 & 53.2 & 65.6& 65.6 & 66.7 & 62.3 & 51.9 \\
        & C2F-TCN~\cite{C2F-TCN} & 72.0 & 56.4 & 61.8 & 57.2 & 45.3 & 66.5 & 48.5 & 53.8 & 48.6 & 40.0 \\
        & LTContext~\cite{LTContext} & 71.4 & 62.5 & 68.1 & 64.3 & 52.1 & 72.2 & 74.2 & 74.6 & 70.0 & 59.6\\
        
        \midrule
        \multirow{5}{*}{\begin{sideways}\textbf{STAS}\end{sideways}}
        & MS-GCN~\cite{MS-GCN} & 65.6 & - & - & - & 43.6 & 64.9 & 52.6 & 52.4 & 48.8 & 39.1 \\
        & ID-GCN+ASRF~\cite{IDT-GCN} & - & - & - & - & - & 67.1& 68.2 & 68.7 & 65.6 & 56.9 \\
        & IDT-GCN~\cite{IDT-GCN} & - & - & - & - & -  & 68.6 & 70.2 & 70.7 & 67.3 & 58.6 \\
        & MTST-GCN~\cite{MTST-GCN} & 73.7 & 58.6 & 63.8 & 59.4 & 47.6 & - & - & - & - & - \\
        & DeST-{\tiny TCN}~\cite{DeST} & 72.6 & 63.7 & 69.7 & 66.7 & 55.8 & 70.5 & 73.8 & 74.0 & 70.7 & 61.8\\
        & DeST-{\tiny Former}~\cite{DeST} & 75.1 & 64.2 & 70.3 & 68.0 & 57.7 & 71.4 & 75.8 & 75.8 & 72.2 & 63.0 \\
        & LaSA~\cite{LaSA} & \underline{75.3} & \textbf{65.7} & \underline{71.6} & \underline{69.0} & \underline{57.9} & \underline{72.6} & \textbf{79.3} & \underline{79.3} & \underline{75.8} & \underline{66.6} \\

        \cmidrule(lr){2-12}
        & \textbf{TRG-Net} & \textbf{76.7} & \underline{65.5} & \textbf{72.3} & \textbf{69.8} & \textbf{59.7} & \textbf{72.6} & \underline{78.5} & \textbf{79.3} & \textbf{76.0} & \textbf{66.7}\\
        \bottomrule
    \end{tabular}
    \label{tab:SOTA2}
\end{table*}

This study quantitatively compares TRG-Net with recent video-based and skeleton-based temporal action segmentation models on the four datasets. To ensure fair comparisons, all methods utilize the same skeleton features as input. As shown in Tables \ref{tab:SOTA1} and \ref{tab:SOTA2}, TRG-Net achieves State-of-The-Art (SOTA) performance across nearly all metrics on the four datasets. The performance improvements are particularly significant on PKU-MMD (X-sub, X-view), and LARa. For the most intuitive metric, \textbf{Acc}, and the most comprehensive metric, \textbf{F1@50}, TRG-Net outperforms LaSA~\cite{LaSA} (the previous SOTA) by +2.1\% and +2.2\% on PKU-MMD (X-sub), +5.6\% and +7.1\% on PKU-MMD (X-view), and +1.4\% and +1.8\% on LARa, substantially exceeding previous SOTA results and other competing methods. These results underscore the effectiveness of text-derived relational graph in TRG-Net for enhanced modeling and supervision, as well as the spatial-aware enhancement processing strategy in TRG-Net.

In addition to its superior performance, TRG-Net maintains reasonable computational efficiency. Its Floating Point Operations Per Second (FLOPs) and parameters are lower than those of the previous SOTA~\cite{LaSA}, and its FLOPs are significantly smaller than those of models such as MS-GCN~\cite{MS-GCN} and MTST-GCN~\cite{MTST-GCN}, demonstrating its good efficiency.

\subsubsection{Qualitative Comparison}
\label{sec:QE}

\begin{figure*}[t]
  \centering
  \begin{subfigure}{0.48\textwidth} 
    \centering
    \includegraphics[width=\linewidth]{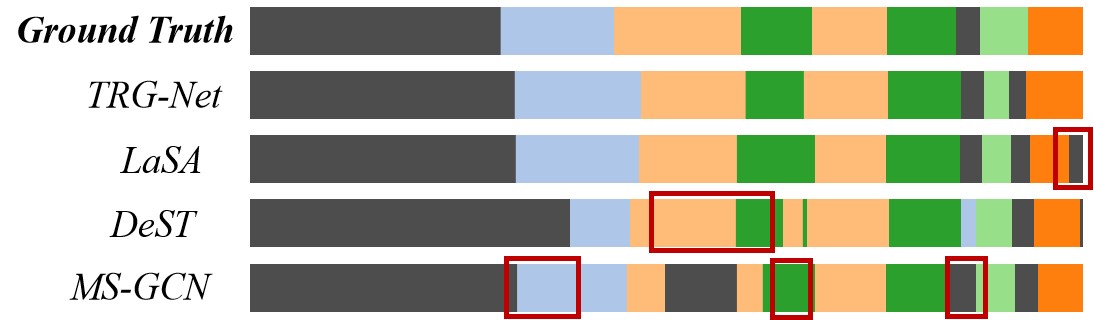} 
    \caption{PKU-MMD (X-sub)}
    \label{fig:6-1}
  \end{subfigure}
  \begin{subfigure}{0.48\textwidth}
    \centering
    \includegraphics[width=\linewidth]{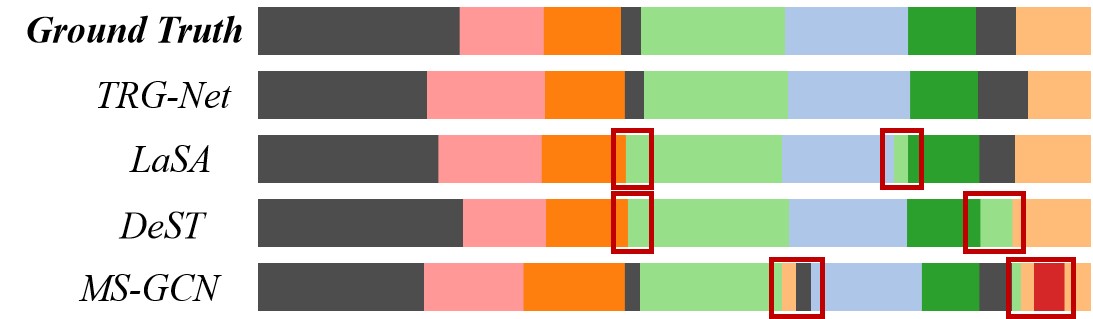} 
    \caption{PKU-MMD (X-view)}
    \label{fig:6-2}
  \end{subfigure}
  
  
  \begin{subfigure}{0.48\textwidth}
    \centering
    \includegraphics[width=\linewidth]{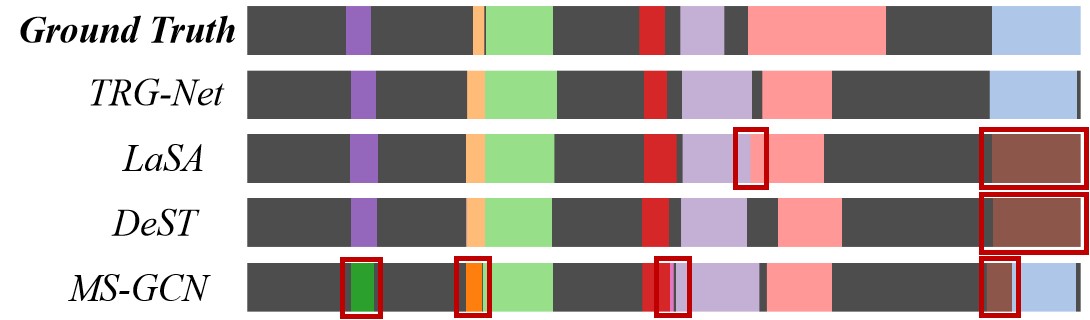} 
    \caption{MCFS-130}
    \label{fig:6-3}
  \end{subfigure}
  \begin{subfigure}{0.48\textwidth}
    \centering
    \includegraphics[width=\linewidth]{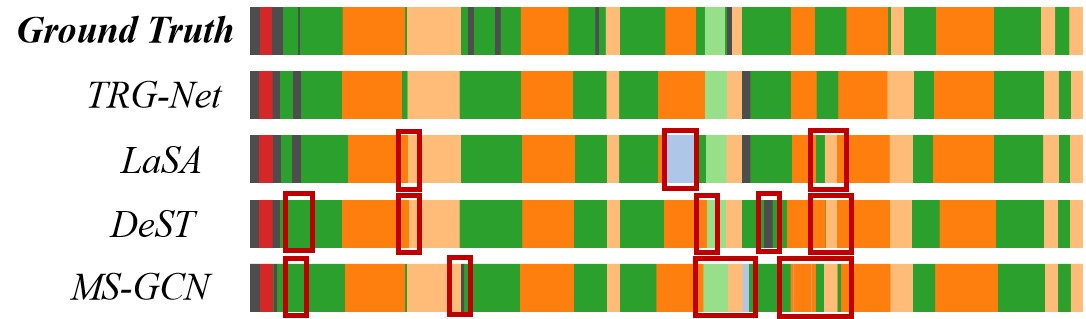} 
    \caption{LARa}
    \label{fig:6-4}
  \end{subfigure}

  \caption{Qualitative results of action segmentation on the PKU-MMD (X-sub, X-view), LARa and MCFS-130 datasets. Different colors represent distinct action classes. Red boxes highlight segmentation errors in other methods compared to TRG-Net, indicating that TRG-Net has fewer segmentation errors and is closer to the ground truth.}
  \label{fig:6}
\end{figure*}

As illustrated in Figure \ref{fig:6}, we present a qualitative comparison of segmentation results between TRG-Net and other competing STAS methods on representative sequences from the four datasets. Compared to TRG-Net, methods such as LaSA~\cite{LaSA}, DeST~\cite{DeST}, and MS-GCN~\cite{MS-GCN} exhibit more frequent classification errors, under-segmentation, and boundary shifts. Additionally, MS-GCN~\cite{MS-GCN} suffers from a higher degree of over-segmentation. TRG-Net, leveraging fine-grained semantic relationships derived from text, achieves more precise modeling and supervision, resulting in fewer segmentation errors and outputs that align more closely with the ground truth.

\subsubsection{Visualization of Representation Space}
\label{sec:VRS}

\begin{figure}[t]
  \centering
  \begin{subfigure}[b]{0.45\textwidth}
    \includegraphics[width=\textwidth]{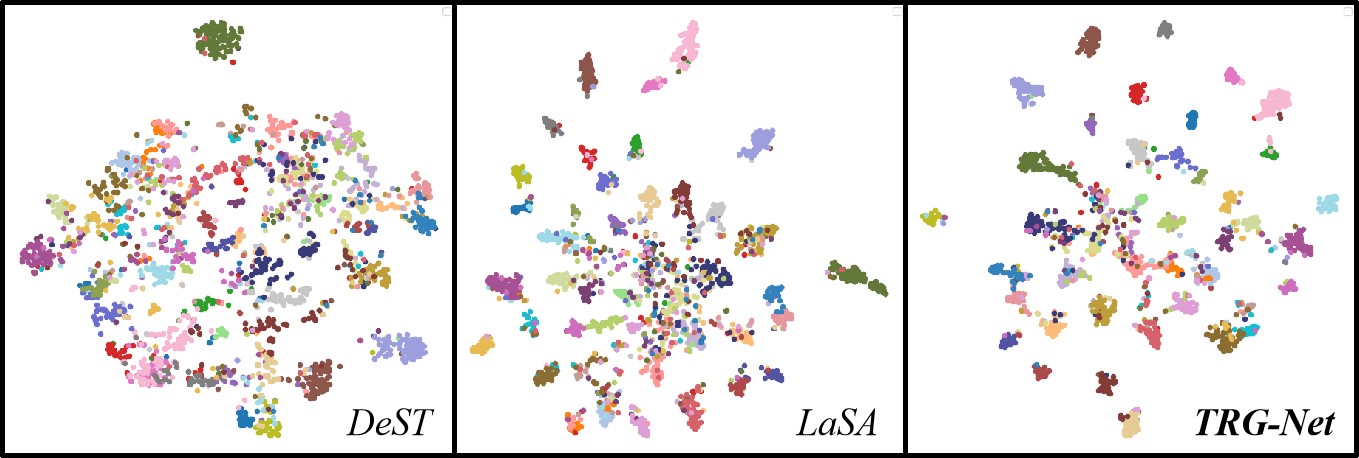}
    \caption{The Representation Space on PKU-MMD(X-subject)}
    \label{fig:7-1}
  \end{subfigure}
  \hfill
  \begin{subfigure}[b]{0.45\textwidth}
    \includegraphics[width=\textwidth]{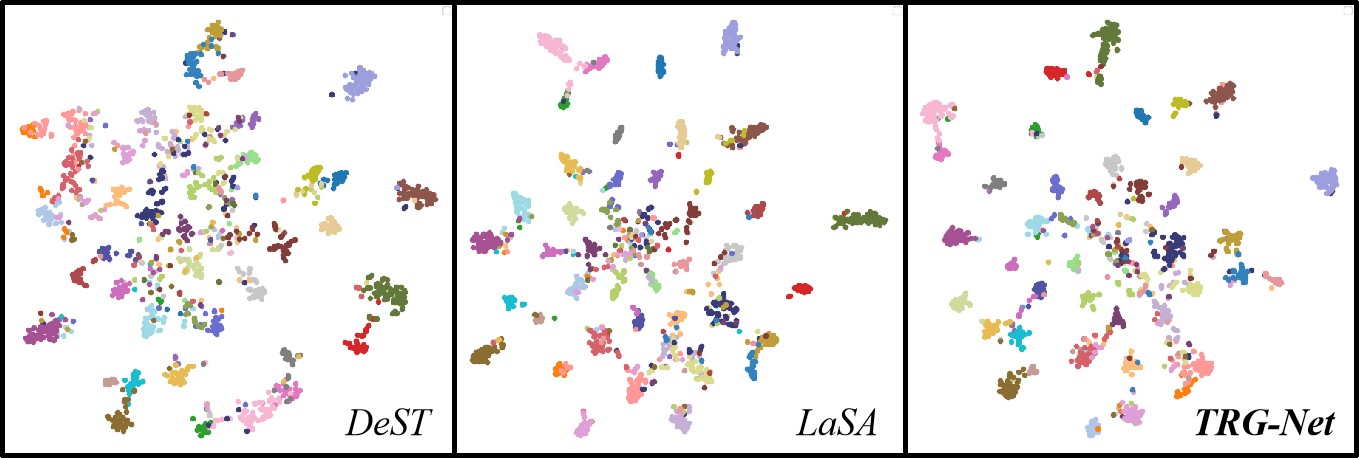}
    \caption{The Representation Space on PKU-MMD (x-view)}
    \label{fig:7-2}
  \end{subfigure}
  
  \caption{Visualization of the representation space on the PKU-MMD (X-sub, X-view) datasets. Each point represents a feature of an action segment, which is colored according to its class label. As observed, compared to others, the ME-ST engenders a more distinctly structured semantic feature space.}

  \label{fig:7}
\end{figure}

We further visualize the representation \(F^t_L\) of TRG-Net alongside those of LaSA~\cite{LaSA} and DeST~\cite{DeST}, the two previous top-performing methods, at the stage following spatio-temporal modeling but preceding the classification and boundary prediction heads Figure \ref{fig:7} visualizes the feature distributions for each action segment. Each point represents a feature \(F_{act} \in \mathbb{R}^C\), derived by slicing \(F^t_L \in \mathbb{R}^{C \times T}\) into segments based on the temporal boundaries in the ground truth and applying temporal average pooling to each segment. It can be observed that LaSA~\cite{LaSA} better aggregates intra-class features and separates inter-class features compared to DeST~\cite{DeST}, owing to its text-assisted action understanding. TRG-Net further enhances these capabilities by modeling and supervising relative semantic relationships through text and incorporating spatial-aware enhancement processing. Consequently, TRG-Net exhibits superior clustering and class discrimination over LaSA~\cite{LaSA}, with a more distinct and structured feature distribution.

\subsection{Effects of Dynamic Spatio-Temporal Fusion Modeling}

\begin{table}[tb]
    \footnotesize
    \centering
    \caption{Effects of text-based adaptive spatial modeling.}
    \begin{tabular}{cc|ccccc}
        \toprule
        Text Graph & Adaptive Graph & Acc & Edit & \multicolumn{3}{c}{F1@\{10, 25, 50\}} \\
        \midrule
        \XSolidBrush & \XSolidBrush & 74.1& 72.9& 78.1& 75.5& 64.9\\
        \Checkmark & \XSolidBrush & 75.0& 73.5& 78.7& 76.0& 65.5\\
        \XSolidBrush & \Checkmark & 74.3& 73.4& 78.5& 75.7& 65.2\\
        \Checkmark & \Checkmark & \textbf{75.4} & \textbf{73.9} & \textbf{79.1} & \textbf{76.4} & \textbf{65.8} \\
        \bottomrule
    \end{tabular}
    \label{tab:TASM}
\end{table}

\subsubsection{Effectiveness of Text-based Adaptive Spatial Modeling}
To demonstrate the effectiveness of text-based adaptive spatial modeling, we evaluate the impact of incorporating the Text-Derived Joint Graph (TJG) and the Dynamic Adaptive Graph (DAG) separately, as shown in Table~\ref{tab:TASM}. The results indicate that adding either TJG or DAG individually improves the performance compared to not using either. Notably, the inclusion of TJG yields a greater improvement, highlighting the powerful role of text-derived priors. Furthermore, the combination of TJG and DAG achieves the best results, demonstrating the effectiveness of text-based adaptive spatial modeling.

\begin{figure}[t]
  \centering
  \begin{subfigure}[b]{0.35\textwidth}
    \includegraphics[width=\textwidth]{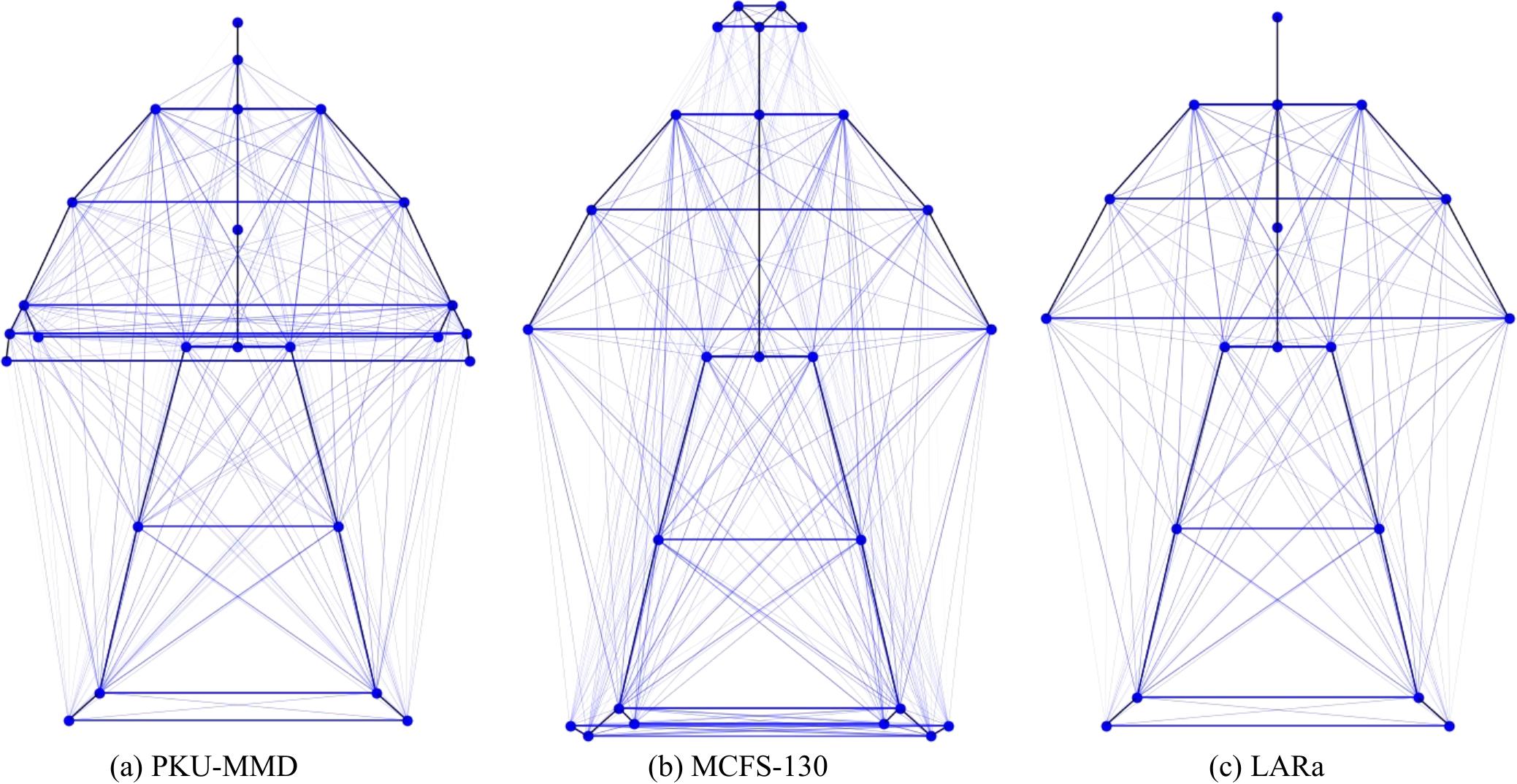}
    \caption{Text-Derived Joint Graphs}
    \label{fig:8-1}
  \end{subfigure}
  \hfill
  \begin{subfigure}[b]{0.40\textwidth}
    \includegraphics[width=\textwidth]{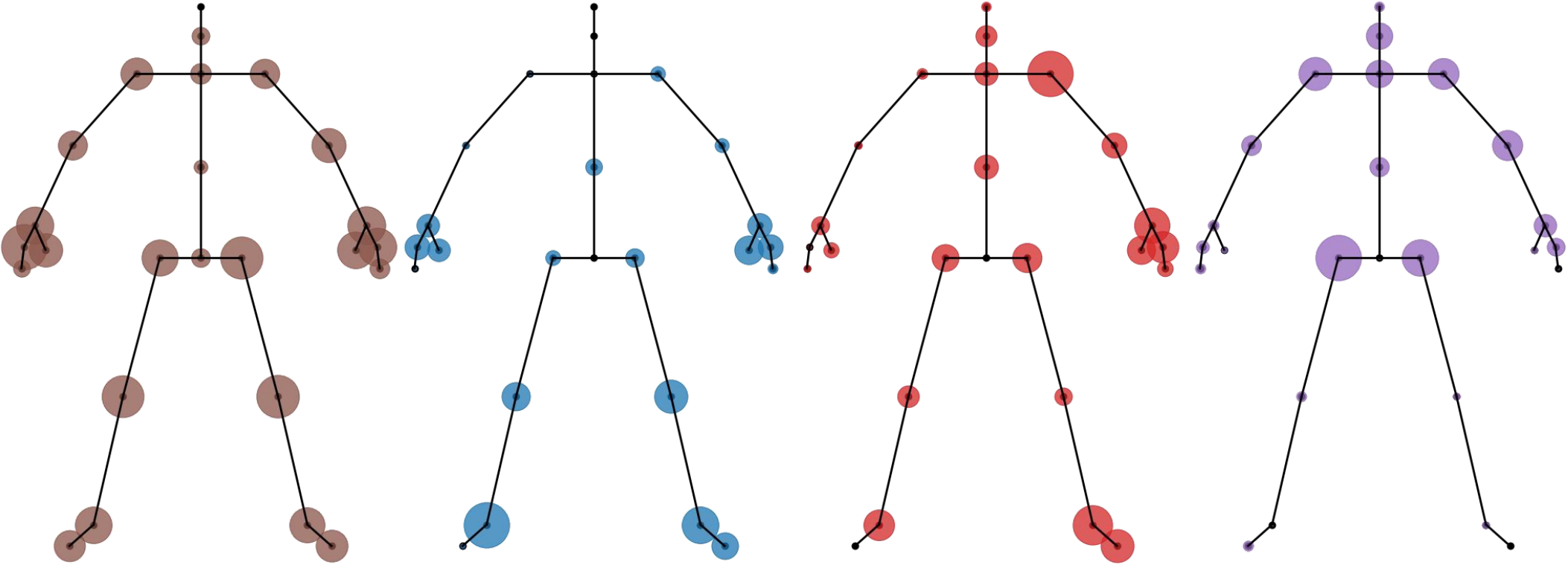}
    \caption{Channel-Adaptive Text-Derived Joint Graphs}
    \label{fig:8-2}
  \end{subfigure}
  \hfill
  \begin{subfigure}[b]{0.46\textwidth}
    \includegraphics[width=\textwidth]{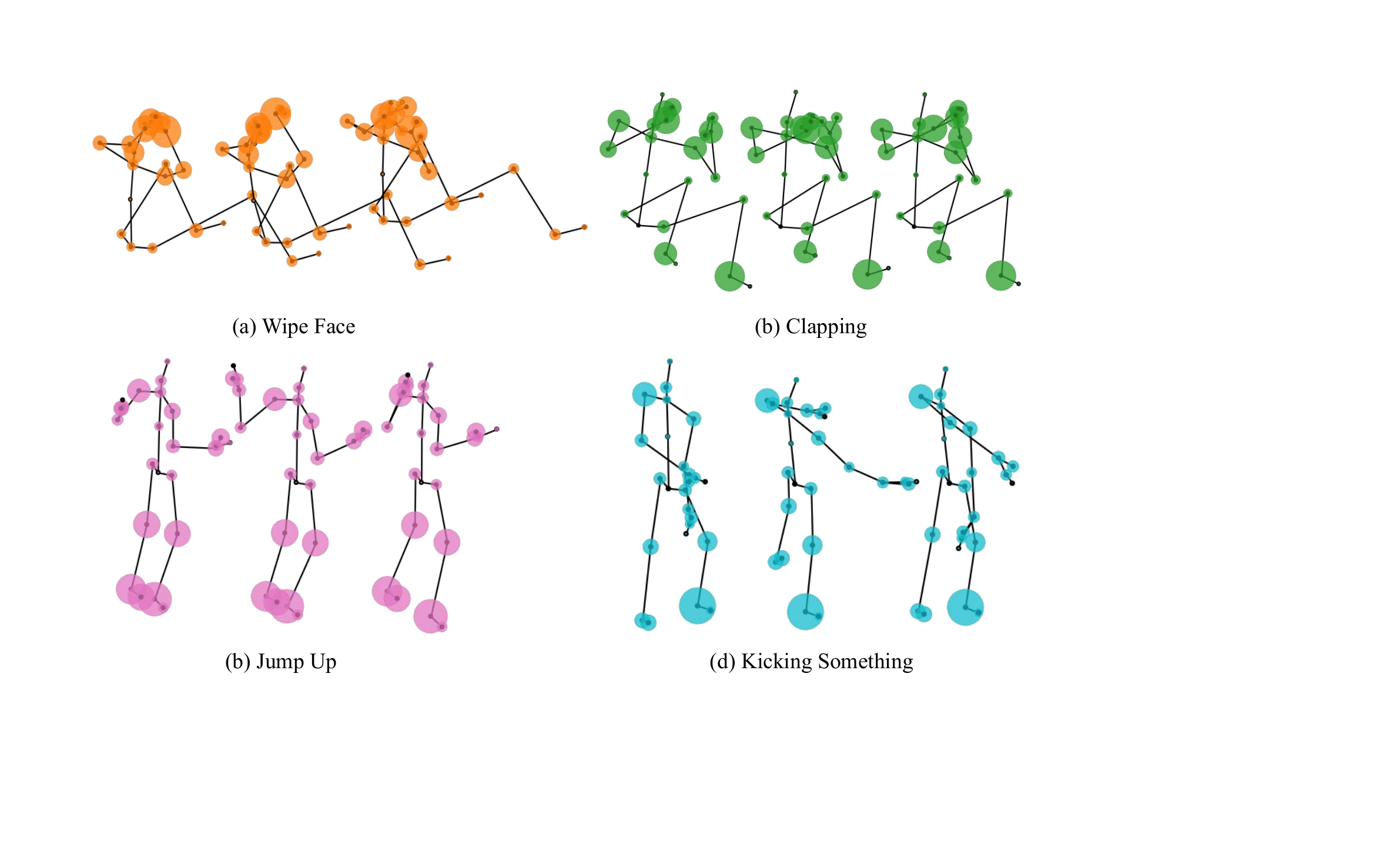}
    \caption{Frame-Adaptive Text-Derived Joint Graphs}
    \label{fig:8-3}
  \end{subfigure}
  
  \caption{Visualization of text-derived joint graph and channel \& frame adaptive graphs. (a) visualizes the Text-Derived Joint Graph (TJG) on skeletons from three datasets, where blue lines indicate the relationships between joints, with thicker and more prominent lines representing higher weights. (b) and (c) illustrate the visualized weights of the Channel-Adaptive Text-Derived Joint Graph (CTJG) and the Frame-Adaptive Text-Derived Joint Graph (FTJG). Larger circles indicate higher aggregated joint weights, and different colors correspond to various channels or action segments. }

  \label{fig:8}
\end{figure}

\subsubsection{Visualization of Text-Derived Joint Graph and Adaptive Adaptations}
To further illustrate the effects of the text-derived joint graph and its dynamic adaptations, we provide visualizations in Figure~\ref{fig:8}. Figure~\ref{fig:8-1} demonstrates the visualization of TJG, which reveals that text-based modeling constructs sufficiently fine-grained semantic joint correlations. The graph shows higher connection weights for symmetric and interactive relationships, aligning well with semantic relevance.
Figure~\ref{fig:8-2} visualizes the aggregated joint weights in the Channel-Adaptive Text-Derived Joint Graph (CTJG), highlighting that different channels indeed focus on distinct joints. Similarly, Figure~\ref{fig:8-3} visualizes the joint weights in the Frame-Adaptive Text-Derived Joint Graph (FTJG), where each action graph is averaged across all frames within its action segment. The results demonstrate that different actions focus on relevant core joints. For instance, “Wipe Face” and “Clapping” emphasize the hands while reducing focus on the legs; “Jump Up” concentrates on the legs with minimal attention to the hands; and “Kicking Something” assigns more weight to one foot while reducing attention elsewhere. These patterns align closely with human intuition regarding joint importance in various actions.

\begin{table}[tb]
    \footnotesize
    \centering
    \caption{Effects of spatio-temporal fusion strategies.}
    \begin{tabular}{l|ccccc}
        \toprule
        Fusion & Acc & Edit & \multicolumn{3}{c}{F1@\{10, 25, 50\}} \\
        \midrule
        None & 74.2& 72.9& 78.5& 75.7& 65.1\\
        Concat \& Conv & \textbf{75.4} & \textbf{73.9} & \textbf{79.1} & \textbf{76.4} & \textbf{65.8} 
\\
        Add & 74.7& 73.0& 78.6& 75.5& 65.4
\\
        Multiplication & 74.4& 73.8& 79.0& 76.3& 65.5
\\
        Attention & 75.3 & 73.6 & 79.0 & 76.2& 65.4\\
        CAM & 74.4& 73.6& 78.8& 76.2& 65.4\\
        \bottomrule
    \end{tabular}
    \label{tab:STFS}
\end{table}

\subsubsection{Effects of Spatio-temporal Fusion Strategies}
To validate the effects of spatio-temporal fusion strategies within DSFM, we compare different fusion methods during temporal modeling, as shown in Table~\ref{tab:STFS}. The results reveal that omitting spatio-temporal fusion yields the lowest performance, emphasizing the necessity of retaining core spatial features during temporal modeling to enhance the capacity. Among the tested strategies, the Concat \& Conv approach performs best. We hypothesize that the Add method is too simplistic to adaptively balance feature weights, while methods such as Multiplication, Attention, and Channel Attention Mechanism (CAM)~\cite{CBAM}, which treat spatial features as attention weights, are less effective for preserving and fusing features. The Concat \& Conv approach, by enabling adaptive feature fusion, emerges as the most effective strategy.

\subsection{Effects of Absolute-Relative Inter-Class Supervision}

\begin{figure}[t]
  \centering
   \includegraphics[width=1\linewidth]{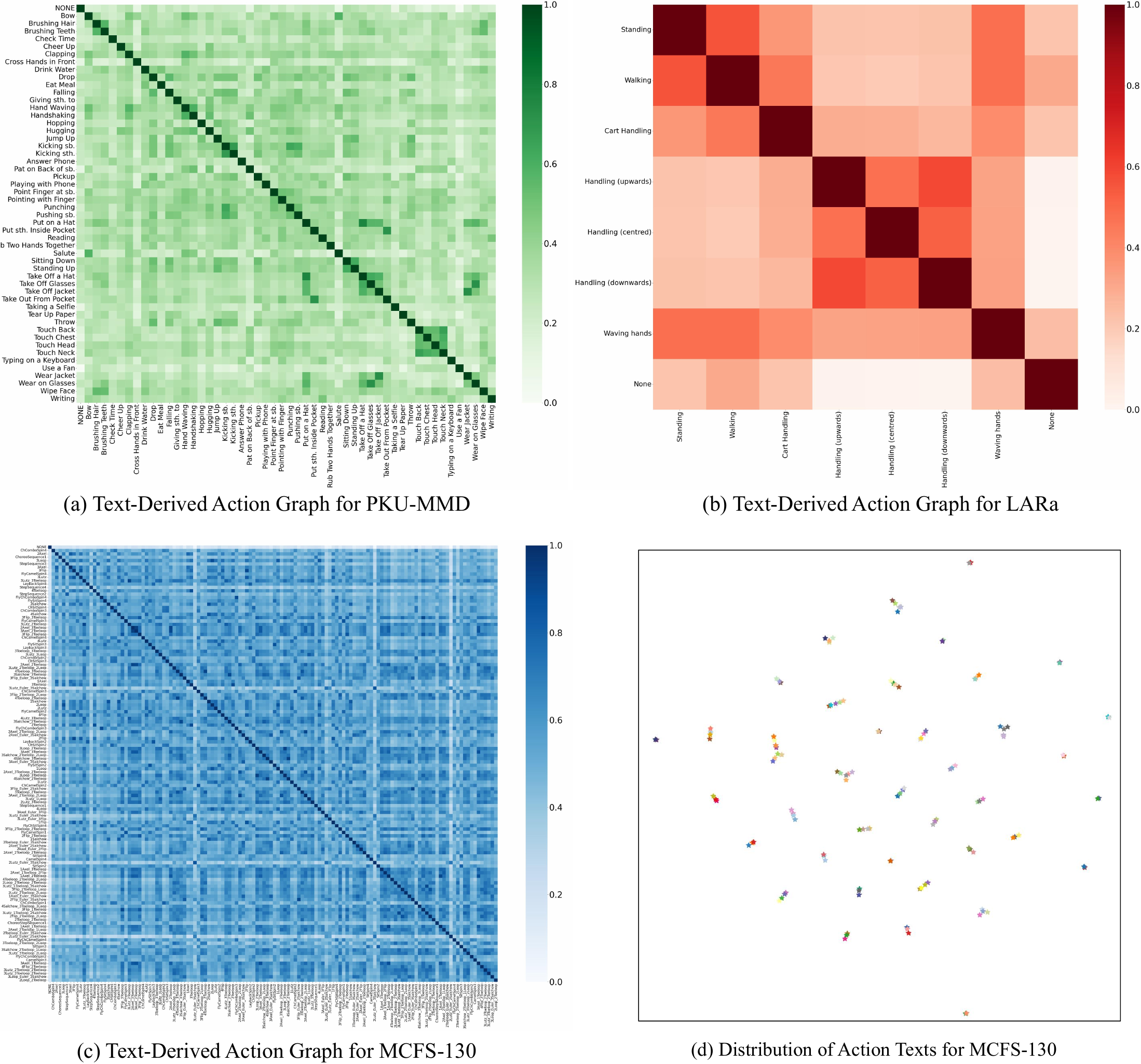}

   \caption{Visualization of text-derived action graph and action text distribution. (a), (b), and (c) depict the Text-Derived Action Graphs (TAG) for three datasets, with each axes representing actions. The color intensity of each cell reflects the relationship between two actions, where darker colors signify higher similarity. (d) visualizes the distribution of action text embeddings for the MCFS-130 dataset, with each star representing an individual action, demonstrating well structured fine-grained relationships among action texts.}
   \label{fig:9}
\end{figure}

\begin{table}[tb]
    \footnotesize
    \centering
    \caption{Effects of absolute \& relative supervision.}
    \begin{tabular}{cc|ccccc}
        \toprule
        Absolute & Relative & Acc & Edit & \multicolumn{3}{c}{F1@\{10, 25, 50\}} \\
        \midrule
        \XSolidBrush & \XSolidBrush & 73.3& 71.5& 76.5& 73.7& 62.6
\\
        \Checkmark & \XSolidBrush & 74.7& 73.7& 78.7& 76.3& 65.3
\\
        \XSolidBrush & \Checkmark & 73.7& 72.3& 77.8& 74.3& 63.3
\\
        \Checkmark & \Checkmark & \textbf{75.4} & \textbf{73.9} & \textbf{79.1} & \textbf{76.4} & \textbf{65.8} \\
        \bottomrule
    \end{tabular}
    \label{tab:ARS}
\end{table}

\subsubsection{Effectiveness of Absolute \& Relative Supervision}
To validate the effectiveness of text-based absolute-relative supervision, we evaluate the impacts of absolute distribution supervision and relative distance supervision, as presented in Table~\ref{tab:ARS}. It can be seen that compared to no text-based supervision, absolute supervision can significantly improve the results, while relative supervision can also slightly improve the performance, reflecting the regularity effect of textual modality on the distribution of visual features. Meanwhile, relative supervision focuses on optimizing the relative relationships between actions, thereby reducing difficulty in fitting the rigid alignment part in absolute supervision. Therefore, combining both can further improve performance.

\subsubsection{Visualization of Text-Derived Action Graph and Semantic Distribution}
To provide an intuitive understanding of the fine-grained relationships between actions in the text-derived action graph and embedding distributions, we visualize the text-derived action graphs for three datasets and the action text embedding distribution for MCFS-130, as shown in Figure~\ref{fig:9}.
The graphs reveal fine-grained relationships between action classes. For instance, in PKU-MMD, ``Put on a hat" is most similar to ``Take out a hat", followed by ``Wear on glasses", ``Wipe face" and ``Wear jacket", then ``Brushing hair" and ``Brushing teeth", while it is least similar to unrelated actions such as ``Jump up". In LARa, the three ``Handling" actions exhibit high similarity among themselves but lower similarity to other actions. In MCFS-130, actions within the same major category, such as various figure-skating movements distinguished only by loop counts, have high similarity, whereas actions across major categories are distributed farther apart. The distribution of action texts for MCFS-130 further confirms this pattern, with subcategories clustering closely within their respective major categories while maintaining well-organized separation between distinct major categories.

\subsection{Effects of Spatial-Aware Enhancement Processing}

\begin{table}[tb]
    \footnotesize
    \centering
    \caption{Effects of occlusion \& rotation data augmentation.}
    \begin{tabular}{ccc|ccccc}
        \toprule
        Occ & Rot & Occ+Rot & Acc & Edit & \multicolumn{3}{c}{F1@\{10, 25, 50\}} \\
        \midrule
        \XSolidBrush & \XSolidBrush & \XSolidBrush & 73.7& 72.2& 77.6& 75& 64.6
\\
        50\% & \XSolidBrush & \XSolidBrush & 74.9& 73.3& 78.8& 75.9& 65.4\\
        \XSolidBrush & 50\% & \XSolidBrush & 74.6& 73.0& 78.5& 75.4& 65.0\\
        33\% & 33\% & \XSolidBrush & \textbf{75.4} & \textbf{73.9} & \textbf{79.1} & \textbf{76.4} & \textbf{65.8} 
\\
        25\% & 25\% & 25\% & 74.7& 73.3& 78& 75.1& 64.5\\
        \bottomrule
    \end{tabular}
    \label{tab:ORDA}
\end{table}

\begin{table}[tb]
    \footnotesize
    \centering
    \caption{Influence of text encoders and feature types.}
    \begin{tabular}{cc|ccccc}
        \toprule
        Encoder & Feature Processing & Acc & Edit & \multicolumn{3}{c}{F1@\{10, 25, 50\}} \\
        \midrule
        \multirow{2}{*}{BERT} & [CLS] token feature  & 74.7& 72.9& 78.6& 75.9& 65.3\\
         & Pooling feature  & 75.4 & 73.9 & \textbf{79.1} & \textbf{76.4} & \textbf{65.8} \\
        \multirow{2}{*}{CLIP} & [CLS] token feature  & 75.1& \textbf{74.1}& 78.7& 76.2& 65.7\\
        & Pooling feature  & \textbf{75.5}& 73.4& 79.0& 76.1& 65.6\\
        \bottomrule
    \end{tabular}
    \label{tab:TEFP}
\end{table}

\subsubsection{Effectiveness of Occlusion \& Rotation Data Augmentation}
To validate spatial-aware enhancement, we evaluate the random joint occlusion and random axial rotation as individual and combined strategies. The results, summarized in Table~\ref{tab:ORDA}, demonstrate that both strategies improve segmentation performance, highlighting their contributions to enhanced spatial generalization.
Further improvement occurs when partial datasets are occluded and partially rotated. However, combining occlusion and rotation on the same portion of the data degrades performance, likely due to excessive information loss and increased variability in the input distribution, which hinders feature learning. As a result, we adopt the strategy of occluding 1/3 of the data and rotating another 1/3.

\subsection{Effects of Text Prompt Priors}

\begin{table}[tb]
    \footnotesize
    \centering
    \caption{Influence of text content types.}
    \begin{tabular}{l|ccccc}
        \toprule
        Text content type & Acc & Edit & \multicolumn{3}{c}{F1@\{10, 25, 50\}} \\
        \midrule
        Action / joint name & 74.6& 73.7& 79.1& 75.6& 65.5\\
        ``This is the action/joint of ..." & 75.0& 73.7& 79.0& 75.8& 65.7\\
        Descriptions by GPT-4 & \textbf{75.4} & \textbf{73.9} & \textbf{79.1} & \textbf{76.4} & \textbf{65.8} \\
        \bottomrule
    \end{tabular}
    \label{tab:TCT}
\end{table}

\begin{table}[t]
    \footnotesize
    \centering
    \caption{Influence of graph construction methods.}
    \begin{tabular}{l|ccccc}
        \toprule
        Relativeness construction & Acc & Edit & \multicolumn{3}{c}{F1@\{10, 25, 50\}} \\
        \midrule
        Pooling \& Subtract & 74.2& 73.9& 79.2& 76& 64.9\\
        Manhattan (L1) Distance & 74.7& 73.9& \textbf{79.5}& 76.2& 65.8\\
        Eucledian (L2) Distance & \textbf{75.4} & 73.9 & 79.1 & \textbf{76.4} & 65.8 \\
        Chebyshev (L$\infty$) Distance & 75.1& 73.6& 79.8& 76.1& 65.6\\
        Cosine Distance & 75.2& 73.9& 79.4& 76.3& \textbf{65.9}\\
        \bottomrule
    \end{tabular}
    \label{tab:RGCM}
\end{table}

\subsubsection{Influence of Text Encoders and Feature Types.}
Table~\ref{tab:TEFP} presents the effect of different text encoders and their corresponding feature types. Both the CLIP~\cite{CLIP} and BERT~\cite{BERT} text encoders achieve strong performance. For CLIP, the two feature types yield nearly identical results. However, for BERT, pooling the embeddings of all words performs better than using only the [CLS] token, consistent with the design of BERT for sentence features. Ultimately, we select BERT as the text encoder and use pooled embeddings as the text features.

\subsubsection{Influence of Text Content Type}
Table~\ref{tab:TCT} summarizes the impact of different text prompt content types. Using simple action/joint names alone results in slightly lower performance due to insufficient semantic richness. Adding prompts such as ``This is the action/joint of..." provides marginal improvement. Furthermore, the use of GPT-4-generated descriptions, which offer rich semantic context, achieves the best results. Therefore, we adopt GPT-4-generated descriptions for text content.

\subsubsection{Influence of Relational Graph Construction Methods}
Table~\ref{tab:RGCM} examines the effects of various methods for constructing relational graphs based on distance calculations. Directly pooling and subtracting two embeddings yields lower performance due to the loss of fine-grained relational information. By contrast, distance-based methods effectively compute similarity and provide comparable results. Among these, the L2 distance is chosen for constructing relational graphs.

\section{Conclusion}
\label{sec:Conclusion}

This study proposes the Text-Derived Relational Graph-Enhanced Network (TRG-Net), which leverages text-derived relational graphs to enhance modeling and supervision. Specifically, the Text-derived Joint Graph (TJG) is employed to facilitate Dynamic Spatio-Temporal Fusion Modeling (DSFM), while the Text-derived Action Graph (TAG) and its embeddings are utilized for Absolute-Relative Inter-Class Supervision (ARIS). Additionally, a Spatial-Aware Enhancement  processing (SAEP) is introduced for to improve spatial generalization. TRG-Net achieves state-of-the-art performance on four challenging datasets. Furthermore, ablation studies validate the effectiveness of the proposed framework from both comparative and interpretability perspectives.

\textbf{Limitations and Future Works.}
While TRG-Net significantly advances skeleton-based action segmentation, certain challenges remain, such as errors in class prediction and boundary prediction shifts, leaving room for improvement. Future research should focus on exploring finer-grained text-guided supervision, integrating text modality adaptively and hierarchically throughout the network to harness richer contextual information. Moreover, leveraging the semantic relationships between joints and actions could guide spatial attention through text-derived priors, further enhancing the capability for action segmentation of the model.

\bibliographystyle{IEEEtran}
\bibliography{references}

\begin{IEEEbiography}[{\includegraphics[width=1in,height=1.25in,clip,keepaspectratio]{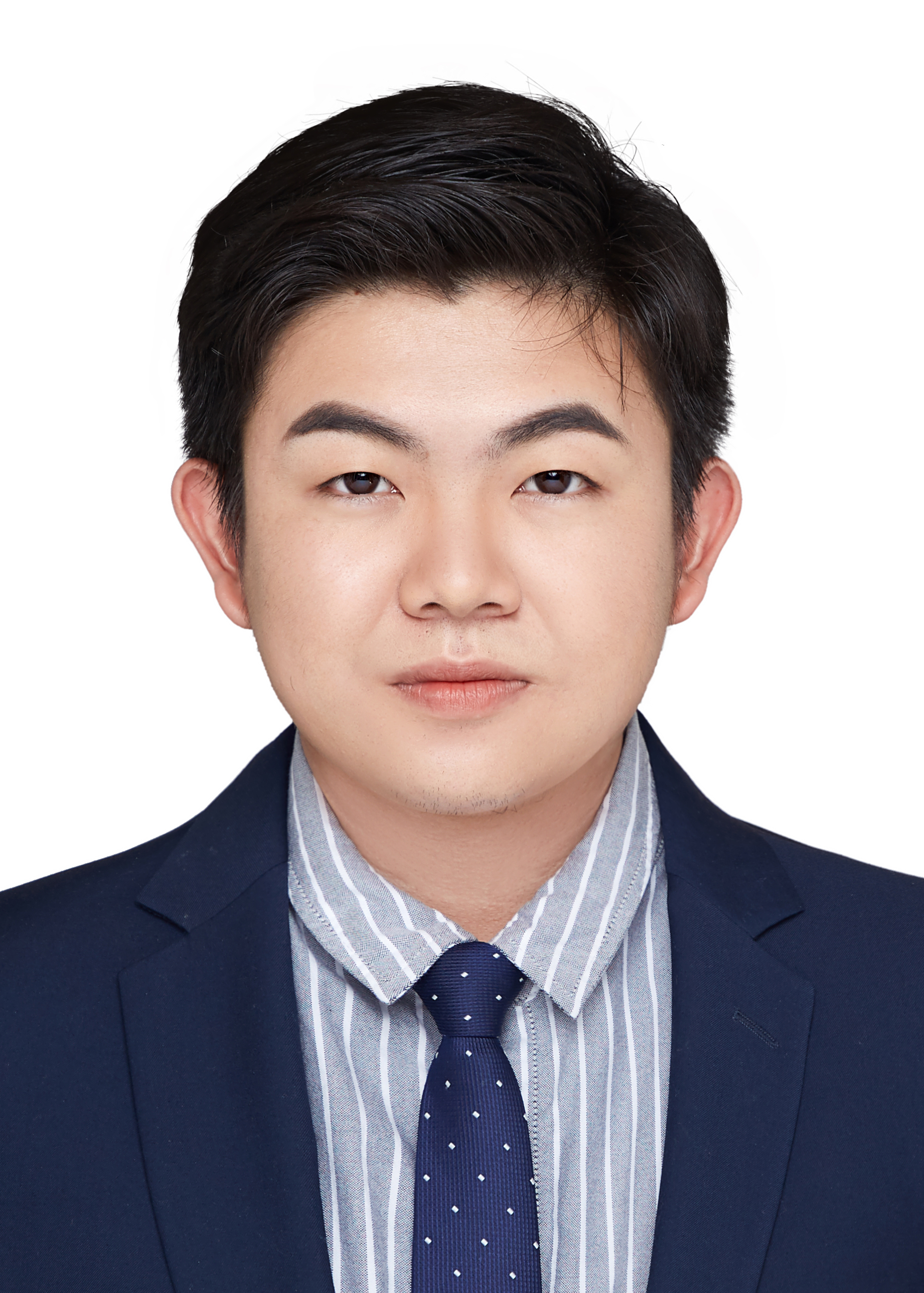}}]{Haoyu Ji}
received his B.E. degree in Mechanical Design, Manufacture, and Automation from Ocean University of China, Qingdao, China, in 2022. He is currently pursuing the Ph.D. with the School of Mechanical Engineering and Automation, Harbin Institute of Technology (Shenzhen), Shenzhen, China. His research interests encompass action segmentation and human behavior understanding in the field of computer vision, with a focus on their applications in autism assessment and intervention.
\end{IEEEbiography}

\begin{IEEEbiography}
[{\includegraphics[width=1in,height=1.25in,clip,keepaspectratio]{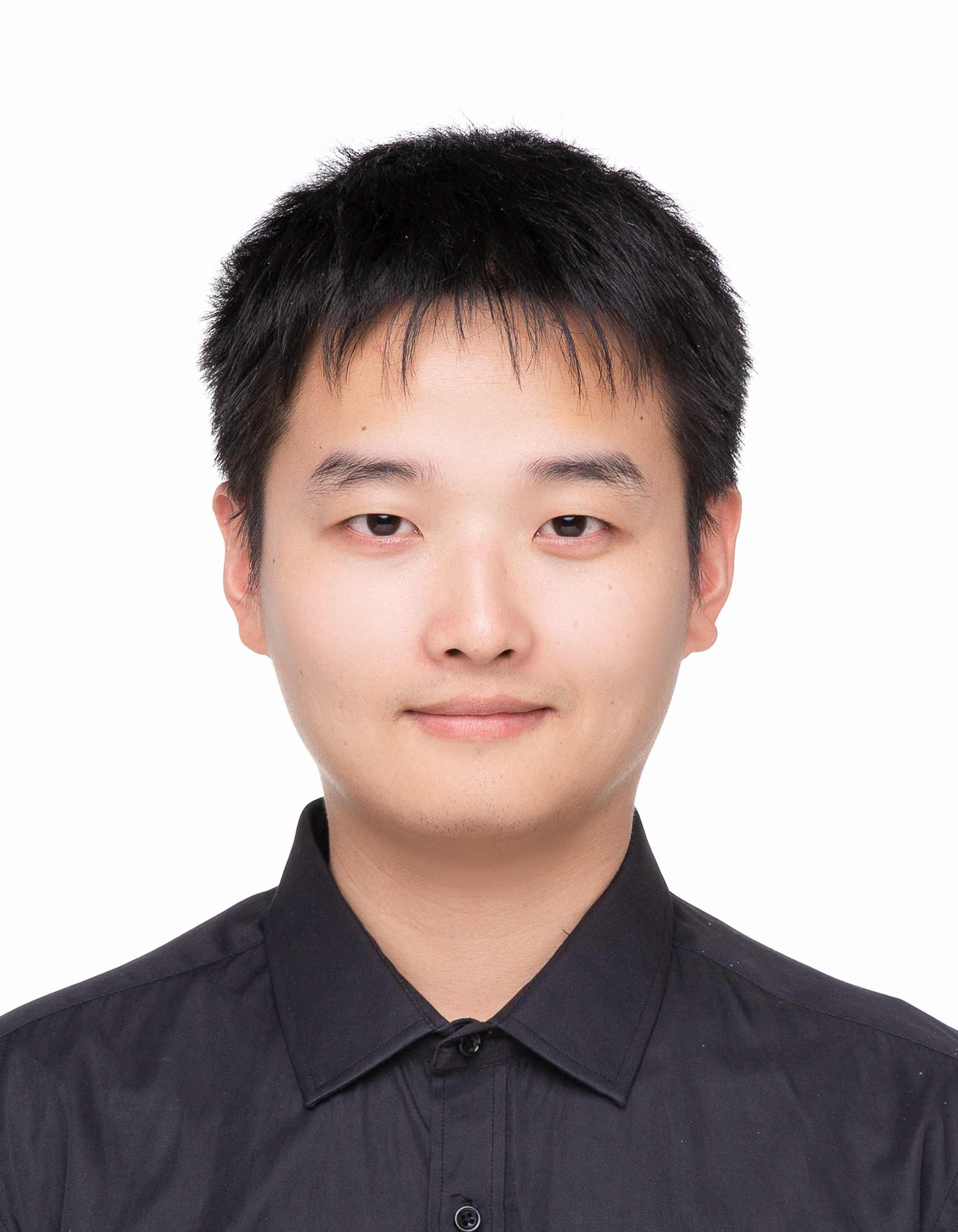}}]{Bowen Chen}
received his B.E. degree in computer science and technology from Hainan University, Haikou, China, in 2016 and the M.S. degree in software engineering from  Zhejiang University of Technology, Hangzhou, China, in 2020. He is currently pursuing the Ph.D. with the School of Mechanical Engineering and Automation, Harbin Institute of Technology (Shenzhen), Shenzhen, China. His research interests include human action understanding and fine-grained interaction analysis.
\end{IEEEbiography}

\begin{IEEEbiography}[{\includegraphics[width=1in,height=1.25in,clip,keepaspectratio]{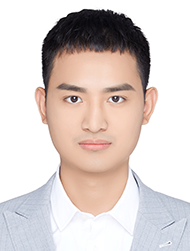}}]{Weihong Ren}
received the B.E. degree in automation and electronic engineering from Qingdao University of Science and Technology, Qingdao, China, in 2013, and the Ph.D. degree from The City University of Hong Kong, Hong Kong, China, and Shenyang Institute of Automation, Chinese Academy of Sciences, Shenyang, China, in 2020. He is an Assistant Professor with the School of Mechanical Engineering and Automation, Harbin
Institute of Technology (Shenzhen), Shenzhen, China. His current research interests include object tracking, action recognition, and deep learning.
\end{IEEEbiography}

\begin{IEEEbiography}
[{\includegraphics[width=1in,height=1.25in,clip,keepaspectratio]{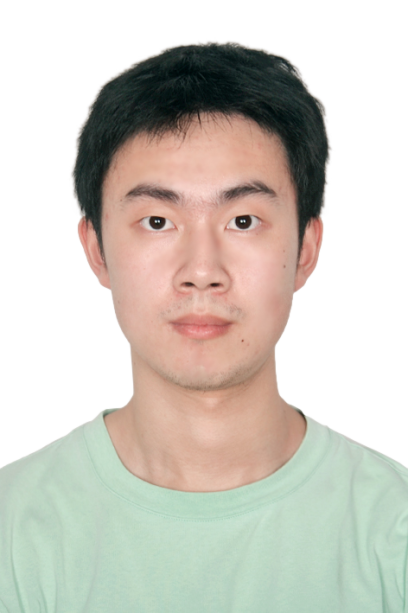}}]{Wenze Huang}
received his B.E. degree in Mechanical Design, Manufacturing, and Automation from Harbin Institute of Technology (Shenzhen), Shenzhen, China. He is currently pursuing his M.S. degree at the School of Mechanical Engineering and Automation, Harbin Institute of Technology, Shenzhen, China. His interests include online action detection and human behavior understanding in the field of computer vision.
\end{IEEEbiography}

\begin{IEEEbiography}
[{\includegraphics[width=1in,height=1.25in,clip,keepaspectratio]{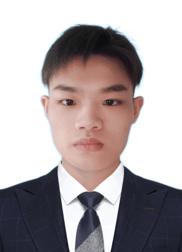}}]{Zhihao Yang} received his B.E. degree in Mechanical Design, Manufacturing, and Automation
from Harbin Institute of Technology (Shenzhen), Shenzhen, China, in 2024. He is currently pursuing his Ph.D. with the School of Mechanical Engineering and Automation, Harbin Institute of Technology, Shenzhen, China. His interests include action analysis and human behavior understanding in the field of computer vision.
\end{IEEEbiography}

\begin{IEEEbiography}[{\includegraphics[width=1in,height=1.25in,clip,keepaspectratio]{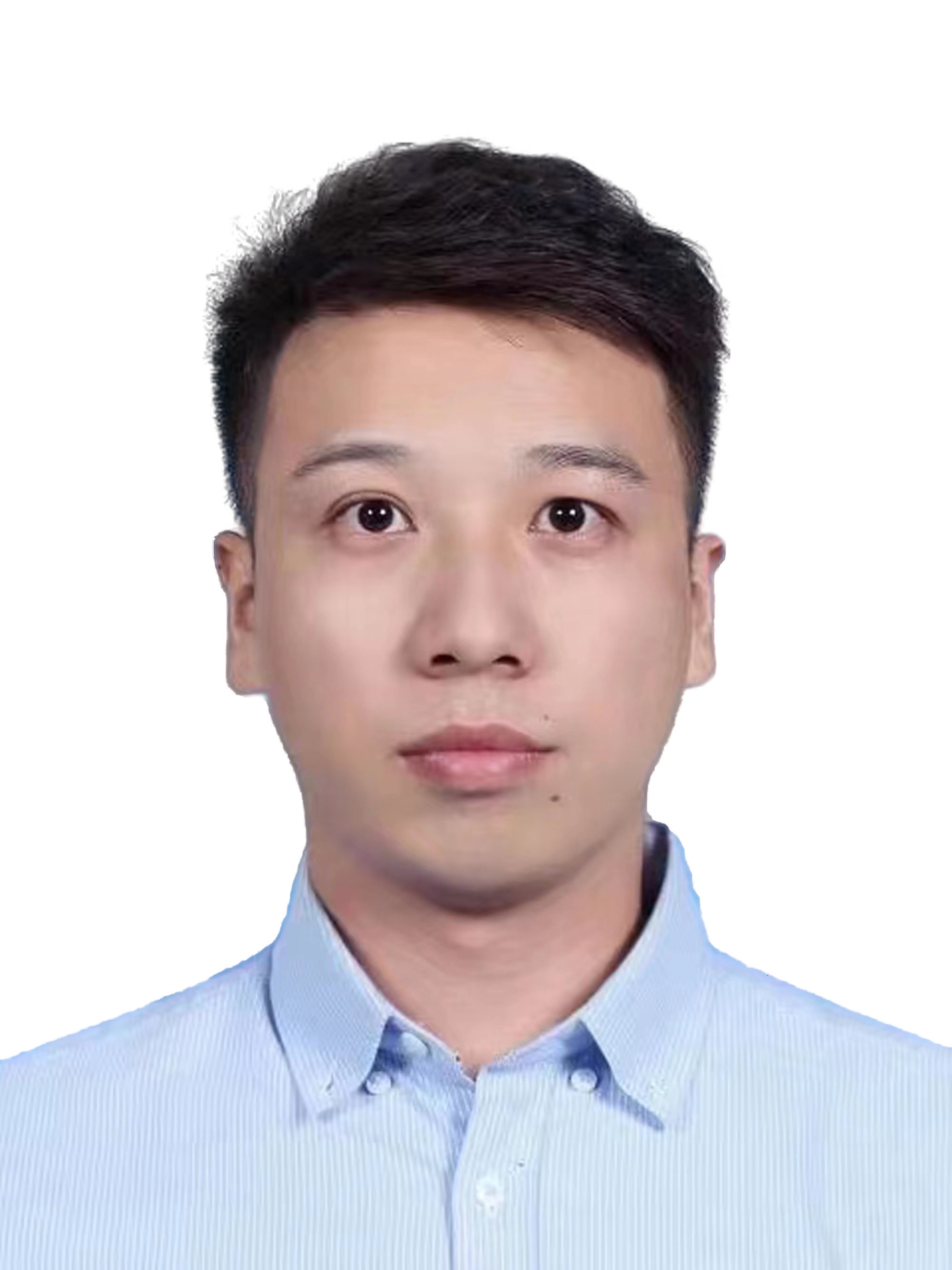}}]{Zhiyong Wang}
(Member, IEEE) received the Ph.D. degree in mechanical engineering from the School of Mechanical Engineering, Shanghai Jiao Tong University, Shanghai, China, in 2023. He is currently an Assistant Professor at Harbin Institute of Technology (Shenzhen), Shenzhen, China. His research interests include human-computer interaction, brain functional response, gaze estimation and their practical applications in medical rehabilitation.
\end{IEEEbiography}

\begin{IEEEbiography}[{\includegraphics[width=1in,height=1.25in,clip,keepaspectratio]{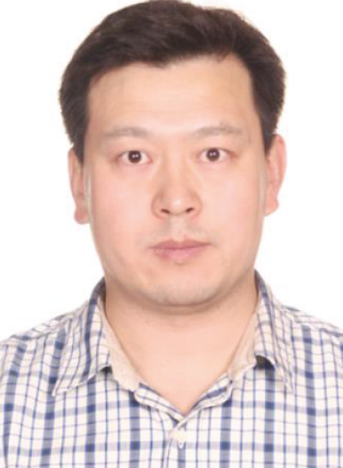}}]{Honghai Liu}
(Fellow, IEEE) received his Ph.D. degree in intelligent robotics from King's College London, London, U.K., in 2003.
He is a Professor at the Harbin Institute of Technology (Shenzhen), Shenzhen, China. 
He is also a Chair Professor of Human–Machine Systems with the University of Portsmouth, Portsmouth, U.K.
His research interests include multisensory data fusion, pattern recognition, intelligent video analytics, intelligent robotics, and their practical
applications.
Prof. Liu is a Co-Editors-in-Chief of the IEEE TRANSACTIONS ON INDUSTRIAL INFORMATICS and an Associate Editor of the IEEE TRANSACTIONS ON CYBERNETICS.
He is the Fellow of the Institution of Engineering and Technology and the Institute of
Electrical and Electronics Engineers. He is also the Member of the Academia Europaea.

\end{IEEEbiography}

\vfill

\appendix
The structure of this supplementary material is as follows: Section \ref{sec:method_supplement} provides further details of methods, including classic spatial and temporal modeling and pre-processing methods. Section \ref{sec:Ablation_Studies} includes additional ablation studies on strategies and hyperparameters, along with their explanations.

\section{Method supplement}
\label{sec:method_supplement}

In this section, we provide additional details on methods including Multi-scale Graph Convolutional Network (GCN) for preliminary spatial modeling, Linear Transformer for global temporal modeling, pre-processing of input data, and other implementation details.

\textbf{Multi-scale GCN for Spatial Modeling.} TRG first adopt a multi-scale GCN inspired by MS-G3D~\cite{MS-G3D} and DeST~\cite{DeST} to capture spatial dependencies among joints. Initially, we define a k-adjacency matrix \(A^{k} \in \{0,1\}^{V \times V}\), connecting joints at a distance of \(k\) as follows:
\begin{equation}
A^{k}_{ij} = 
\begin{cases} 
1, & \text{if } d(\alpha_i,\alpha_j) = k \text{ or } i = j,\\
0, & \text{otherwise},
\end{cases}
\end{equation}
where \(d(\alpha_i,\alpha_j)\) represents the shortest distance (the number of edges) between joints \(\alpha_i\) and \(\alpha_j\). Subsequently, we concatenate all adjacency matrices from 0 to the maximum scale \(K\) to form the multi-scale adjacency matrix \(A^{MS} \in \{0,1\}^{V \times KV}\):
\begin{small}
\begin{equation}
A^{MS} = [(\tilde{D}^1)^{-\frac{1}{2}} A^{1} (\tilde{D}^1)^{-\frac{1}{2}}] \oplus \cdot \cdot \cdot \oplus [(\tilde{D}^K)^{-\frac{1}{2}} A^{K} (\tilde{D}^K)^{-\frac{1}{2}})],
\end{equation}
\end{small}
where \(\oplus\) represents concatenation along the second dimension. \(\tilde{D}^k\) is a diagonal matrix normalizing \(A^k\), with \(\tilde{D}_{ii}^k=\sum_{j}(A_{ij}^k)+ \alpha\), where \(\alpha= 0.001\) is introduced to prevent empty rows in\(A^k\). The scale \(K\) is set to 13. Given the input sequence \(X \in \mathbb{R}^{C_0 \times T \times V}\), the multi-scale spatial features \(F^{g}\) obtained after multi-scale GCN can be represented as:
\begin{equation}
F^{g} = \mathrm{ReLU}[\mathrm{reshape}((A^{MS} + B) X) W_s],
\end{equation}
 where \(B \in \mathbb{R}^{V \times KV}\) is a trainable matrix which can adaptively learn the relationships between joints. \(\mathrm{reshape}(\cdot)\) denotes the operation that transforms the feature from \(\mathbb{R}^{C_0 \times T \times KV}\)  to \(\mathbb{R}^{KC_0 \times T \times V}\). \(W_s \in \mathbb{R}^{1 \times 1 \times KC_0 \times C}\) represents the convolution operator for channel adjustment.

\textbf{Linear Transformer for Temporal Modeling.}
Unlike Temporal Convolution Networks (TCN), which captures features within a fixed temporal receptive field, the attention mechanism in Transformer can adaptively model dependencies among all frames. However, due to the high dimensionality of the temporal features of action segmentation sequences, the memory requirements for global attention in conventional transformers are prohibitively high, leading to the necessity of using only local window attention~\cite{Asformer, UVAST}. In order to leverage global attention, we adopt the Linear Transformer~\cite{Linformer1, Linformer2} used in~\cite{DeST, LaSA} as our temporal modeling method, which reduces the complexity of attention from \(O(n^2)\) to \(O(n)\), enabling global temporal modeling. Formally, the Linear Transformer layer can be computed as follows:
\begin{equation}
\label{con:linformer}
F^{t}_l = \mathrm{ReLU}\left[\phi(Q^t_l) (\phi(K^t_l)^T V^t_l) \cdot W^t_l + F^{st}_l \right].
\end{equation}
Here, the fused features \(F^{st}_l \in \mathbb{R}^{C \times T} \) which integrate upper-layer temporal and spatial core features are used as the input to the Linear Transformer. \(Q^t_l\), \(K^t_l\), and \(V^t_l \in \mathbb{R}^{C_3 \times T}\) are obtained from \(F^{st}_l\) via linear layers \(W_{Qt}\), \(W_{Kt}\), and \(W_{Vt} \in \mathbb{R}^{C \times C_3}\), respectively. The linear layer \(W_t \in \mathbb{R}^{C_3 \times C}\) is used for channel adjustment, while \(\phi(\cdot)\) represents the sigmoid activation function. The channel \(C_3\) is set to 16. Finally, the Linear Transformer outputs the feature \(F^{t}_l \in \mathbb{R}^{C \times T}\) after temporal modeling of the current layer.

\textbf{Data Pre-processing.}
We adopt the same approach as~\cite{DeST, LaSA} for input data processing. Specifically, for MCFS~\cite{MCFS}, we directly use its 2-axis positions to derive 2-channel relative positions. Regarding PKU-MMD~\cite{PKU-MMD}, we employ its 3-axis positions to calculate 6-channel relative positions and displacements. As for LARa~\cite{LARA}, we leverage its 3-axis positions and orientations to compute 12-channel relative positions and orientations, along with their corresponding relative displacements.

\textbf{Other Implementation Details.}
For the Linear Transformer, each layer employs 4 attention heads. The 10 layers of Linear Transformer in the spatio-temporal backbone utilize self-attention as shown in Equation~\ref{con:linformer}, while the 10 layers in the class prediction branch at each stage use cross-attention, where the Query and Key are derived from the current layer features and the Value from the final features of the previous stage. Additionally, a dropout rate of 0.5 is applied to both the backbone and each stage of the two branches for temporal modeling (linear transformer or dilated TCN) to mitigate overfitting.

\section{Ablation Studies}
\label{sec:Ablation_Studies}
This section explores the impact of various strategies and hyperparameters on model performance, including the influence of dynamic channel \& frame adaptations, loss function type in the text-based relative supervision, supervision in the refinement class prediction branch, occlusion strategies and ratios, rotation strategies and angles, text-derived graph normalization methods, and the number of refinement branches stages.

\begin{table}[tb]
    \footnotesize
    \centering
    \caption{Effects of dynamic channel \& frame adaptations.}
    \begin{tabular}{cc|ccccc}
        \toprule
        Channel & Frame & Acc & Edit & \multicolumn{3}{c}{F1@\{10, 25, 50\}} \\
        \midrule
        \XSolidBrush & \XSolidBrush & 75.0& 73.5& 78.7& 76.0& 65.4\\
        \Checkmark & \XSolidBrush & 74.9& 73.6& 78.8& 76.3& 65.5\\
        \XSolidBrush & \Checkmark & 75.0& 73.9& 79.0& 76.4& 65.7\\
        \Checkmark & \Checkmark & \textbf{75.4} & \textbf{73.9} & \textbf{79.1} & \textbf{76.4} & \textbf{65.8} \\
        \bottomrule
    \end{tabular}
    \label{tab:DCFA}
\end{table}

\subsubsection{Effects of Dynamic Channel \& Frame Adaptations}
To evaluate the impact of channel- and frame-level dynamic adaptations, we examine their respective effects on the adaptive refinement of the text-derived joint graph. As shown in Table~\ref{tab:DCFA}, both adaptations enhance model performance, with frame-level adaptation contributing the most significant improvements. This is attributed to its ability to focus on key motion joints within temporal sequences, aiding in action differentiation. When both channel- and frame-level adaptations are applied, the model achieves even better results, underscoring the effectiveness of dynamic adaptations.

\begin{table}[tb]
    \footnotesize
    \centering
    \caption{Effect of relative supervised loss functions.}
    \begin{tabular}{l|ccccc}
        \toprule
        Loss Type & Acc & Edit & \multicolumn{3}{c}{F1@\{10, 25, 50\}} \\
        \midrule
        MSE Loss & 74.5& 72.9& 79.0& 76.2& 65.3\\
        L1 Loss & 74.8 & 73.8 & 78.7 & 75.9 & 65.5 \\
        KL Divergence & \textbf{75.4} & \textbf{73.9} & \textbf{79.1} & \textbf{76.4} & \textbf{65.8} \\
        \bottomrule
    \end{tabular}
    \label{tab:RSLF}
\end{table}

\subsubsection{Effect of Relative Supervised Loss Function Types}
We investigated the selection of loss functions for aligning the relationship of action features to those in text-derived action graph in relative supervision, as shown in Table~\ref{tab:RSLF}. Among the loss functions tested, KL divergence outperforms MSE and L1 losses. This superiority arises from its suitability for aligning two relationship distributions. KL divergence aligns distributions of relations by emphasizing the structural details of the target distribution, making it more effective for capturing inter-class relationships, whereas MSE and L1 focus on point-wise errors, limiting their ability to leverage global distribution information.


\begin{table}[tb]
    \footnotesize
    \centering
    \caption{Effect of supervision in class prediction branch.}
    \begin{tabular}{l|ccccc}
        \toprule
        Branch Supervision & Acc & Edit & \multicolumn{3}{c}{F1@\{10, 25, 50\}} \\
        \midrule
        None & \textbf{75.4} & \textbf{73.9} & \textbf{79.1} & \textbf{76.4} & \textbf{65.8} \\
        Absolute & 74.8& 74.5& 79.2& 76.4& 65.1\\
        Relative & 74.4& 73.5& 78.6& 75.4& 64.9\\
        Absolute + Relative & 74.9& 73.2& 78.2& 75.3& 64.7\\
        \bottomrule
    \end{tabular}
    \label{tab:SCPB}
\end{table}

\subsubsection{Effect of Supervision in Refinement Class Prediction Branch}
Currently, absolute-relative inter-class supervision is applied only to feature representations before the primary classification and boundary heads, not to the subsequent refinement branches. To evaluate the impact of applying this supervision to the refinement class prediction branch, we conduct additional experiments, as summarized in Table~\ref{tab:SCPB}.
The results indicate that applying absolute, relative, or both types of supervision to the refinement branch reduces model performance. This is likely because the refinement branch operates on features closer to class probabilities rather than feature representations, leading to a mismatch between the supervision type and the target features. Consequently, this misalignment causes optimization conflicts, degrading the overall model performance.

\begin{table}[tb]
    \footnotesize
    \centering
    \caption{Effects of occlusion strategies and ratios.}
    \begin{tabular}{cc|ccccc}
        \toprule
        Strategy & Ratio  & Acc & Edit & \multicolumn{3}{c}{F1@\{10, 25, 50\}} \\
        \midrule
        \multirow{3}{*}{\makecell{Sequence-\\wise}} & 25\%  & 74.6& 73.4& 78.7& 75.6& 65.4\\
         & 50\%  & 74.7 & 73.5 & 79.1 & 76.1 & 65.5 
\\
         & 0-50\%  & \textbf{75.4} & \textbf{73.9} & \textbf{79.1} & \textbf{76.4} & \textbf{65.8} 
\\
        \multirow{3}{*}{\makecell{Frame-\\wise}} & 25\%  & 75.0& 73.6& 78.6& 76.3& 65.7
\\
        & 50\%  & 74.5& 73.4& 78.7& 75.1& 65.6
\\
        & 0-50\%  & 74.0& 73.4& 78.7& 75.8& 65.3\\
        \bottomrule
    \end{tabular}
    \label{tab:OSR}
\end{table}

\subsubsection{Effect of Occlusion Strategies and Ratios}
We further investigate the impact of occlusion strategies and ratios, as shown in Table~\ref{tab:OSR}. Occlusion strategies are categorized into sequence-level (identical joints occluded across all frames) and frame-level (different joints occluded per frame). Occlusion ratios include 25\%, 50\%, and a random 0–50\%.
The results indicate minor differences among the strategies and ratios, with sequence-level occlusion outperforming frame-level occlusion. This superiority may arise from the reduced noise in sequence-level occlusion, as frame-level occlusion introduces random variability that is harder to model. The best performance is achieved with sequence-level occlusion using a random 0–50\% ratio, which is selected as the final approach.

\begin{table}[tb]
    \footnotesize
    \centering
    \def\degree{${}^{\circ}$}
    \caption{Effects of rotation strategies and angles.}
    \begin{tabular}{cc|ccccc}
        \toprule
        Main axis & Others & Acc & Edit & \multicolumn{3}{c}{F1@\{10, 25, 50\}} \\
        \midrule
        0-30 \degree& \XSolidBrush  & \textbf{75.5}& 73.8& 78.9& 76.2& 65.6\\
        0-360 \degree& \XSolidBrush & 75.4 & 73.9 & \textbf{79.1} & \textbf{76.4} & \textbf{65.8} \\
        0-360 \degree& 0-30 \degree& 75.1& \textbf{74.2}& 79.1& 76.3& 65.8\\
        0-30 \degree& 0-30 \degree& 75.3& 73.6& 78.8& 76.2& 65.7\\
        0-360 \degree& 0-360 \degree& 73.3& 73.3& 78.1& 74.9& 62.8\\
        \bottomrule
    \end{tabular}
    \label{tab:RSA}
\end{table}

\subsubsection{Effect of Rotation Strategies and Angles}
The effects of rotation strategies and angles are analyzed in Table~\ref{tab:RSA}. Here, Main axis refers to the gravitational (axial) axis. Results show that randomly rotating around the main axis consistently improves performance, with rotations of 0–360° yielding the greatest benefit. This aligns with the ``isotropy" of action orientations in real-world scenarios, emphasizing the importance of main-axis rotation for spatial generalization.
For other axes, small-angle rotations have negligible impact, but large-angle rotations significantly degrade performance. This likely results from excessive feature alterations, which hinder effective model fitting. Consequently, we adopt random 0–360° main-axis rotations as the optimal strategy.

\begin{table}[tb]
    \footnotesize
    \centering
    \caption{Influence of graph normalization methods.}
    \begin{tabular}{l|ccccc}
        \toprule
        Graph normalization & Acc & Edit & \multicolumn{3}{c}{F1@\{10, 25, 50\}} \\
        \midrule
        Min-Max normalization & \textbf{75.4} & \textbf{73.9} & 79.1 & \textbf{76.4} & 65.8 
\\
        Z-Score normalization & 75.4& 73.8& \textbf{79.2}& 76.3& \textbf{65.9}\\
        Sigmoid function & 75.2& 73.7& 79.1& 76.3& 65.8\\
        \bottomrule
    \end{tabular}
    \label{tab:GNM}
\end{table}

\subsubsection{Influence of Graph Normalization Methods}
Table~\ref{tab:GNM} compares the impact of different graph normalization methods. Normalization aids spatial modeling and relative supervision by refining the relative relationships within the graph. While different methods introduce subtle variations in graph structure, the robust fitting capability of the model minimizes performance differences. We select Min-Max to scale graph values between 0 and 1 for simplicity and consistency.

\begin{table}[tb]
    \footnotesize
    \centering
    \caption{Influence of number of branch stages.}
    \begin{tabular}{cc|ccccc}
        \toprule
        Class branch & Boundary branch & Acc & Edit & \multicolumn{3}{c}{F1@\{10, 25, 50\}} \\
        \midrule
        \(S^c= 1\) & \(S^b= 1\) & 75.1& 73.9& 78.5& 75.8& 64.7
\\
        \(S^c= 1\) & \(S^b= 2\) & \textbf{75.4} & \textbf{73.9} & \textbf{79.1} & \textbf{76.4} & \textbf{65.8} 
\\
        \(S^c= 2\) & \(S^b= 2\) & 74.4& 73.2& 78.7& 75.5& 65.4\\
        \(S^c= 2\) & \(S^b= 3\) & 74.2& 73.0 & 78.7& 75.8& 64.9\\
        \(S^c= 3\) & \(S^b= 3\) & 73.5& 71.9& 77.8& 74.6& 63.8\\
        \bottomrule
    \end{tabular}
    \label{tab:NBS}
\end{table}

\subsubsection{Influence of Number of Branch Stages}
The effect of the number of refinement branch stages is evaluated in Table~\ref{tab:NBS}. For the class prediction branch, a single stage achieves the best performance, as additional stages lead to overfitting, with linear transformers sufficiently refining predictions in one stage. Conversely, for the boundary regression branch, two stages are optimal, as as two stages of TCN layers effectively capture boundary regression requirements. Thus, we set the class prediction branch stages \(S^c= 1\) and the boundary regression branch stages \(S^b= 2\), balancing efficiency with near-optimal performance.

\end{document}